# Conditional Accelerated Lazy Stochastic Gradient Descent

**Guanghui Lan**      GEORGE.LAN@ISYE.GATECH.EDU
*School of Industrial & Systems Engineering*
*Georgia Institute of Technology*
*Atlanta, GA 30332, USA*

**Sebastian Pokutta**      SEBASTIAN.POKUTTA@ISYE.GATECH.EDU
*School of Industrial & Systems Engineering*
*Georgia Institute of Technology*
*Atlanta, GA 30332, USA*

**Yi Zhou**      YIZHOU@GATECH.EDU
*School of Industrial & Systems Engineering*
*Georgia Institute of Technology*
*Atlanta, GA 30332, USA*

**Daniel Zink**      DANIEL.ZINK@GATECH.EDU
*School of Industrial & Systems Engineering*
*Georgia Institute of Technology*
*Atlanta, GA 30332, USA*

## Abstract

In this work we introduce a conditional accelerated lazy stochastic gradient descent algorithm with optimal number of calls to a stochastic first-order oracle and convergence rate $O(\frac{1}{\varepsilon^2})$ improving over the projection-free *Online Frank-Wolfe* based stochastic gradient descent of Hazan and Kale (2012) with convergence rate $O(\frac{1}{\varepsilon^4})$.

## 1. Introduction

The conditional gradient method (also known as: Frank-Wolfe algorithm) proposed in Frank and Wolfe (1956), gained much popularity in recent years due to its simple projection-free scheme and fast practical convergence rates. We consider the basic convex programming (CP) problem

$$f^* := \min_{x \in X} f(x), \tag{1.1}$$

where $X \subseteq \mathbb{R}^n$ is a closed convex set and $f : X \to \mathbb{R}$ is a smooth convex function such that $\exists L > 0$,

$$\|f'(x) - f'(y)\|_* \leq L\|x - y\|, \quad \forall x, y \in X. \tag{1.2}$$

The classic conditional gradient (CG) method solves (1.1) iteratively by minimizing a series of linear approximations of $f$ over the feasible set $X$. More specifically, given $x_{k-1} \in X$ at the $k$-th iteration, it updates $x_k$ according to the following steps:

1) Call the first-order (FO) oracle to compute $(f(x_{k-1}), f'(x_{k-1}))$ and set $p_k = f'(x_{k-1})$.

2) Call the linear optimization (LO) oracle to compute

$$y_k \in \operatorname{argmin}_{x \in X} \langle p_k, x \rangle. \tag{1.3}$$

3) Set $x_k = (1 - \lambda_k)x_{k-1} + \lambda_k y_k$ for some $\lambda_k \in [0, 1]$.

Compared to most other first-order methods, such as, gradient descent algorithms and accelerated gradient algorithms Nesterov (1983, 2004), the CG method is computationally cheaper in some cases, since it only requires the solution of a linear optimization subproblem (1.3) rather than an often costly projection onto the feasible region $X$.

There has been extensive and fruitful research on the general class of linear-optimization-based convex programming (LCP) methods (which covers the CG method and its variants) and their applications in machine learning (e.g., Ahipasaoglu and Todd (2013); Bach et al. (2012); Beck and Teboulle (2004); Cox et al. (2013); Clarkson (2010); Freund and Grigas (2013); Hazan (2008); Harchaoui et al. (2012); Jaggi (2011, 2013); Jaggi and Sulovský (2010); Luss and Teboulle (2013); Shen et al. (2012); Hazan and Kale (2012); Lan (2013); Lan and Zhou (2014); Braun et al. (2016)). It should be noted that even the computational cost for LO oracle to solve the linear optimization subproblem (1.3) is high for some complex feasible regions. Recently, several approaches have been considered to address this issue. Jaggi (2013) demonstrated practical speed up for the CG method by approximately solving (1.3). Braun et al. (2016) proposed a class of modified CG methods, namely the lazy conditional gradient (LCG) algorithms, which call a weak separation oracle rather than solving the linear subproblem (1.3) in the classical CG method. In fact, the weak separation oracle is computationally more efficient than approximate minimization used in Jaggi (2013), at the expense of not providing any guarantee for function value improvement with respect to (1.3). Furthermore, as shown in Jaggi (2013) and Lan (2013), the total number of iterations for the LCP methods to find an $\epsilon$-solution of (1.1) (i.e., a point $\bar{x} \in X$, s.t. $f(\bar{x}) - f^* \leq \epsilon$) cannot be smaller than $\mathcal{O}(1/\epsilon)$, which is not improvable even when the objective function $f$ is strongly convex. Improved complexity results can only be obtained under stronger assumptions on the LO oracle or the feasible set (see, e.g., Garber and Hazan (2013); Lan (2013)). However, the $\mathcal{O}(1/\epsilon)$ bound does not preclude the existence of more efficient LCP algorithms for solving (1.1). Lan and Zhou (2014) proposed a class of conditional gradient sliding methods (CGS), which significantly improve the complexity bounds in terms of the number of gradient evaluations while maintaining optimal complexity bounds for the LO oracle calls required by the LCP methods.

Inspired by Braun et al. (2016) and Lan and Zhou (2014), in this paper we focus on a class of modified LCP methods that require only improving solutions for a certain separation problem rather than solving the linear optimization subproblem (1.3) explicitly through LO oracle calls while simultaneously minimizing the number of gradient evaluations when performing weak separation over the feasible set $X$. At first these two objectives seem to be incompatible as Braun et al. (2016) gives up the dual guarantee to simplify the oracle, while the dual guarantee of CG iterations is at the core of the analysis in Lan and Zhou (2014). We overcome this impasse by carefully modifying both techniques.

It should be mentioned that Hazan and Kale (2012) proposed the online Frank-Wolfe (OFW) algorithm, which obtains $\mathcal{O}(1/\epsilon^4)$ rate of convergence for stochastic problems. Indeed, if we consider the objective function $f(x) := \mathbb{E}[F(x, \xi)]$ for stochastic optimization, the OFW method can be applied to solve (1.1) by viewing the iteratively observed function $f_t$ as the current realization of the true objective function $f$, i.e., $f_t(\cdot) = F(\cdot, \xi_t)$. *Without re-evaluating the (sub)gradients at the updated points*, the OFW obtains $\mathcal{O}(T^{-1/4})$ bound for any (smooth or non-smooth) objective functions (see Theorem 4.4 in Hazan and Kale (2012)), which implies $\mathcal{O}(1/\epsilon^4)$ rate of convergence in terms of the number of (sub)gradient evaluations for stochastic optimization. However, we can show that our proposed algorithm obtains $\mathcal{O}(1/\epsilon^2)$ rate of convergence for stochastic problems, which is much better than the convergence rate of the OFW method. We would like to stress that the stochastic optimization bound in Hazan and Kale (2012, Theorem 4.1) which gives a guarantee of $\mathcal{O}(1/\epsilon^2)$, requires to re-evaluate all gradients at the current iterate and as such the number of gradient evaluations required grows quadratically in $t$.

Moreover, Hazan and Luo (2016) proposed two methods for solving the special case of Problem (1.1) of the form

$$\min_{x \in X} f(x) = \min_{x \in X} \frac{1}{m} \sum_{i=1}^{m} f_i(x),$$

which allows for a potentially smaller number of SFO evaluations than $\mathcal{O}(1/\varepsilon^2)$, the lower bound for the general problem. The two methods Stochastic Variance-Reduced Frank-Wolfe (SVRF) and Stochastic Variance-Reduced Conditional Gradient Sliding (STORC) are obtained by applying the variance reduction idea of Johnson and Zhang (2013) and Mahdavi et al. (2013) to the CG method and the Stochastic CGS method respectively. Both algorithms however need a certain number of exact (or full) gradient evaluations leading to a potentially undesirable dependence on the number of examples $m$. This idea of applying variance reduction technique can also be adapted to our proposed method CALSGD.

**Contributions**

Our main contributions can be briefly summarized as follows. We consider stochastic smooth optimization, where we have only access to unbiased estimators of the gradients of $f$ via a stochastic first-order (SFO) oracle. By incorporating a modified LCG procedure (Braun et al., 2016) into a modified CGS method (Lan and Zhou, 2014) we obtain a new conditional accelerated lazy stochastic gradient descent algorithm (CALSGD) and we show that the number of calls to the weak separation oracle can be optimally bounded by $\mathcal{O}(1/\epsilon)$, while the optimal bound of $\mathcal{O}(1/\epsilon^2)$ on the total number of calls to the SFO oracle can be maintained. In addition, if the exact gradients of $f$ can be accessed by an FO oracle, the latter bound can be significantly improved to $\mathcal{O}(1/\sqrt{\epsilon})$. In order to achieve the above we will present a modified lazy conditional gradient method, and show that the total number of iterations (or calls to the weak separation oracle) performed by it can be bounded by $\mathcal{O}(1/\epsilon)$ under a stronger termination criterion, i.e., the primal-dual gap function.

We also consider strongly convex and smooth functions and show that without enforcing any stronger assumptions on the weak separation oracle or the feasible set $X$, the total number of calls to the FO (resp., SFO) oracle can be optimally bounded by $\mathcal{O}(\log 1/\epsilon)$ (resp., $\mathcal{O}(1/\epsilon)$) for variants of the proposed method to solve deterministic (resp., stochastic) strongly convex and smooth problems. Furthermore, we also generalize the proposed algorithms to solve an important class of non-smooth convex programming problems with a saddle point structure. By adaptively approximating the original non-smooth problem via a class of smooth functions, we are able to show that the deterministic version of CALSGD can obtain an $\epsilon$-solution within $\mathcal{O}(1/\epsilon)$ number of linear operator evaluations and $\mathcal{O}(1/\epsilon^2)$ number of calls to the weak separation oracle, respectively. The former bound will increase to $\mathcal{O}(1/\epsilon^2)$ for non-smooth stochastic optimization. We also extend CALSGD to solve a general non-smooth stochastic problem. Comparing to OFW proposed by Hazan and Kale (2012), which requires $\mathcal{O}(1/\epsilon^4)$ number of calls to the SFO oracle to obtain a stochastic $\epsilon$-solution, CALSGD improves it to an optimal $\mathcal{O}(1/\epsilon^2)$ while obtaining a comparable $\mathcal{O}(1/\epsilon^4)$ number of calls to the weak separation oracle.

Finally, we demonstrate practical speed ups of CALSGD through preliminary numerical experiments for the video co-localization problem, the structured regression problem and quadratic optimization over the standard spectrahedron; an extensive study is beyond the scope of this paper and left for future work. In all cases we report a substantial improvements in performance.

### 1.1 Notation and terminology

Let $X \subseteq \mathbb{R}^n$ be a convex compact set, and $\|\cdot\|_X$ be the norm associated with the inner product in $\mathbb{R}^n$. For the sake of simplicity, we often skip the subscript in the norm $\|\cdot\|_X$. We define the diameter of the set $X$ as

$$D_X \equiv D_{X, \|\cdot\|} := \max_{x,y \in X} \|x - y\|. \tag{1.4}$$

For a given norm $\|\cdot\|$, we denote its conjugate by $\|s\|_* = \max_{\|x\| \leq 1} \langle s, x \rangle$. For a linear operator $A : \mathbb{R}^n \to \mathbb{R}^m$, we use $\|A\|$ to denote its operator norm defined as $\|A\| := \max_{\|x\| \leq 1} \|Ax\|$. Let $f : X \to \mathbb{R}$ be a convex function, we denote its linear approximation at $x$ by

$$l_f(x; y) := f(x) + \langle f'(x), y - x \rangle. \tag{1.5}$$

Clearly, if $f$ satisfies (1.2), then

$$f(y) \leq l_f(x; y) + \tfrac{L}{2}\|y - x\|^2, \quad \forall x, y \in X. \tag{1.6}$$

Notice that the constant $L$ in (1.2) and (1.6) depends on $\|\cdot\|$. Moreover, we say $f$ is *smooth with curvature* at most $C$, if

$$f(y) \leq l_f(x; y) + \tfrac{C}{2}, \quad \forall \ x, y \in X. \tag{1.7}$$

It is clear that if $X$ is bounded, we have $C \leq LD_X^2$. In the following we also use $\mathbb{R}_{++}$ to denote the set of strictly positive reals.

### 2. Conditional Accelerated Lazy Stochastic Gradient Descent

We now present a new method for stochastic gradient descent that is based on the stochastic conditional gradient sliding (SCGS) method and the parameter-free lazy conditional gradient (LCG) procedure from Section 2.2, which we refer to as the *Conditional Accelerated Lazy Stochastic Gradient Descent (CALSGD)* method.

We consider the stochastic optimization problem:
$$f^* := \min_{x \in X}\{f(x) = \mathbb{E}_\xi[F(x,\xi)]\}, \tag{2.1}$$
where $f(x)$ is a smooth convex function satisfying (1.2).

### 2.1 The algorithm

Throughout this section, we assume that there exists a stochastic first-order (SFO) oracle, which for a search point $z_k \in X$ outputs a stochastic gradient $F'(z_k, \xi_k)$, s.t.
$$\mathbb{E}\left[F'(z_k, \xi_k)\right] = f'(z_k), \tag{2.2}$$
$$\mathbb{E}\left[\|F'(z_k, \xi_k) - f'(z_k)\|_*^2\right] \le \sigma^2. \tag{2.3}$$
If $\sigma = 0$, the stochastic gradient $F'(z_k, \xi_k)$ is the exact gradient at point $z_k$, i.e., $F'(z_k, \xi_k) = f'(z_k)$.

Our algorithmic framework is inspired by the SCGS method by Lan and Zhou (2014). However, instead of applying the classic CG method to solve the projection subproblem appearing in the accelerated gradient (AG) method, the CALSGD method utilizes a modified parameter-free LCG algorithm (see Section 2.2) to approximately solve the subproblem $\psi(x)$ defined in (2.9) and skips the computations of the stochastic gradient $F'(z, \xi)$ from time to time when performing weak separation over the feasible region $X$. The main advantages of our method are that it does not solve a traditional projection problem and achieves the optimal bounds on the number of calls to the SFO and LOsep$_X$ oracles (see Oracle 1 in Subsection 2.2) for solving problem (1.1)-(2.1). To the authors' best knowledge, no such algorithms have been developed before in the literature; we present the algorithm below in Algorithm 1.

---
**Algorithm 1** Conditional Accelerated Lazy Stochastic Gradient Descent (CALSGD)

**Input:** Initial point $x_0 \in X$, iteration limit $N$, and weak separation oracle accuracy $\alpha \ge 1$. Let $\beta_k \in \mathbb{R}_{++}, \gamma_k \in [0,1]$, and $\eta_k \in \mathbb{R}_+$, $k = 1, 2, \ldots$, be given and set $y_0 = x_0$.
**for** $k = 1, 2, \ldots, N$ **do**

$$z_k = (1 - \gamma_k)y_{k-1} + \gamma_k x_{k-1}, \tag{2.4}$$
$$g_k = \tfrac{1}{B_k}\sum_{j=1}^{B_k} F'(z_k, \xi_{k,j}), \tag{2.5}$$
$$x_k = \text{LCG}(g_k, \beta_k, x_{k-1}, \alpha, \eta_k), \tag{2.6}$$
$$y_k = (1 - \gamma_k)y_{k-1} + \gamma_k x_k, \tag{2.7}$$

where $F'(z_k, \xi_{k,j})$, $j = 1, \ldots, B_k$, are stochastic gradients computed by the SFO at $z_k$.
**end for**
**Output:** $y_N$.

---

We hasten to make some observations about the CALSGD method. Firstly, we apply mini-batches to estimate the gradient at point $z_k$, where the parameter $\{B_k\}$ denotes the batch sizes used to compute $g_k$. It can be easily seen from (2.2), (2.3), and (2.5) that
$$\mathbb{E}[g_k - f'(z_k)] = 0 \text{ and } \mathbb{E}[\|g_k - f'(z_k)\|_*^2] \le \tfrac{\sigma^2}{B_k}, \tag{2.8}$$

and hence $g_k$ is an unbiased estimator of $f'(z_k)$. In fact, letting $S_{B_k} = \sum_{j=1}^{B_k}(F'(z_k, \xi_{k,j}) - f'(z_k))$, from (2.2) and (2.3), by induction, we have

$$\begin{aligned}
\mathbb{E}\left[\|S_{B_k}\|_*^2\right] &= \mathbb{E}\left[\|S_{B_k-1} + F'(z_k, \xi_{k,B_k}) - f'(z_k)\|_*^2\right] \\
&= \mathbb{E}\left[\|S_{B_k-1}\|_*^2 + \|F'(z_k, \xi_{k,B_k}) - f'(z_k)\|_*^2 + 2\langle S_{B_k-1}, F'(z_k, \xi_{k,B_k}) - f'(z_k)\rangle\right] \\
&= \mathbb{E}\left[\|S_{B_k-1}\|_*^2\right] + \mathbb{E}\left[\|F'(z_k, \xi_{k,B_k}) - f'(z_k)\|_*^2\right] \\
&= \sum_{j=1}^{B_k} \mathbb{E}\left[\|F'(z_k, \xi_{k,j}) - f'(z_k)\|_*^2\right] \leq B_k \sigma^2,
\end{aligned}$$

which together with the fact that $g_k - f'(z_k) = \frac{1}{B_k}\sum_{j=1}^{B_k}[F'(z_k, \xi_{k,j}) - f'(z_k)] = \frac{1}{B_k}S_{B_k}$, implies the second relationship in (2.8).

Secondly, in view of the SCGS method in Lan and Zhou (2014), $x_k$ obtained in (2.6) should be an approximate solution to the gradient sliding subproblem

$$\min_{x \in X}\left\{\psi_k(x) := \langle g_k, x\rangle + \frac{\beta_k}{2}\|x - x_{k-1}\|^2\right\}, \tag{2.9}$$

such that for some $\eta_k \geq 0$ we have

$$\langle \psi_k'(x_k), x_k - x\rangle = \langle g_k + \beta_k(x_k - x_{k-1}), x_k - x\rangle \leq \eta_k, \tag{2.10}$$

for all $x \in X$. If we solve the subproblem (2.9) exactly (i.e., $\eta_k = 0$), then CALSGD will reduce to the accelerated stochastic approximation method by Lan (2009, 2012). However, by employing the LCG procedure (see Procedure 1 in Subsection 2.2), we only need to use a weak separation oracle, but still maintaining the optimal bounds on stochastic first-order oracle as in Lan (2009, 2012); Lan and Zhou (2014).

Thirdly, observe that the CALSGD method so far is conceptual only as we have not yet specified the LCG procedure and the parameters $\{B_k\}$, $\{\beta_k\}$, $\{\gamma_k\}$, and $\{\eta_k\}$. We will come back to this issue after introducing the LCG procedure and establishing its main convergence properties.

### 2.2 The parameter-free lazy conditional gradient procedure

The classical CG method is a well-known projection-free algorithm, which requires only the solution of a linear optimization subproblem (1.3) rather than the projection over $X$ per iteration. Therefore, it has computational advantages over many other first-order methods when projection over $X$ is costly. The LCG procedure presented in this subsection, a modification of the vanilla LCG method in Braun et al. (2016), goes several steps further than CG and even the vanilla LCG method. Firstly, it replaces LO oracle by a weaker separation oracle LOsep, which is no harder than linear optimization and often much simpler. Secondly, it uses a stronger termination criterion, the Frank-Wolfe gap (cf. (2.11)), than vanilla LCG method. Finally, it maintains the same order of convergence rate as the CG and the vanilla LCG method.

We present the LOsep oracle in Oracle 1 below. Observe that the oracle has two output modes. In particular, Oracle 1 first verifies whether there exists an improving point $y \in P$ with the required guarantee and if so it outputs this point, which we refer it as a *positive call*. If no such point exists the oracle certifies this by providing the maximizer $y$, which then also provides a new duality gap. We refer to this case as a *negative call*. The computational

---

**Oracle 1** Weak Separation Oracle LOsep$_P(c, x, \Phi, \alpha)$

**Input:** $c \in \mathbb{R}^n$ linear objective, $x \in P$ point, $\alpha \geq 1$ accuracy, $\Phi > 0$ objective value;
**Output:** $y \in P$ vertex with either (1) $c^T(x-y) > \Phi/\alpha$, or (2) $y = \text{argmax}_{y \in P} c^T(x-z) \leq \Phi$.

---

advantages of this oracle are that it can reuse previously seen solutions $y$ if they satisfy the improvement condition and even if the LO oracle has to be called, the optimization can be terminated early once the improvement condition is satisfied. Finally, the parameter $\alpha$ allows to only approximately satisfy the improvement condition making separation even easier; in our applications we set the parameter $\alpha$ slightly larger than 1.

We present the LCG procedure based on Braun et al. (2016) below. We adapted the parameter-free version to remove any dependence on hard to estimate parameters. For any smooth convex function $\phi$, we define its *duality gap* as

$$gap_{\phi,X}(x) \equiv gap_\phi(x) := \max_{y \in X} \nabla\phi(x)^T(x-y). \tag{2.11}$$

Clearly, by convexity the duality gap is an upper bound on $f(x) - f(x^*)$. Given any accuracy parameter $\eta \geq 0$, the LCG procedure solves $\min_{x \in X} \phi(x)$ approximately with accuracy $\eta$, i.e., it outputs a point $\bar{u} \in X$, s.t. $gap_\phi(\bar{u}) \leq \eta$.

---

**Procedure 1** Parameter-free Lazy Conditional Gradients (LCG) procedure

**Input:** access to gradients of smooth convex function $\phi$, $u_1 \in X$ vertex, LOsep$_X$ weak linear separation oracle, accuracy $\alpha \geq 1$, duality gap bound $\eta$
**Output:** $\bar{u} \in X$ with bounded duality gap, i.e., $gap_\phi(\bar{u}) \leq \eta$
1: $\Phi_0 \leftarrow \max_{u \in X} \nabla\phi(u_1)^T(u_1 - u)$
2: **for** $t = 1$ to $T - 1$ **do**
3: $\quad v_t \leftarrow \text{LOsep}_X(\nabla\phi(u_t), x_t, \Phi_{t-1}, \alpha)$
4: $\quad$ **if not** $\nabla\phi(u_t)^T(u_t - v_t) > \Phi_{t-1}/\alpha$ **then**
5: $\quad\quad$ **if** $\Phi_{t-1} = \eta$ **then**
6: $\quad\quad\quad$ **return** $\bar{u} = u_t$
7: $\quad\quad$ **end if**
8: $\quad\quad \Phi_t \leftarrow \max\left\{\frac{\Phi_{t-1}}{2}, \eta\right\}$ {Update $\Phi_t$}
9: $\quad$ **end if**
10: $\quad \lambda_t \leftarrow \text{argmin}\, \phi((1-\lambda_t)u_t + \lambda_t v_t)$
11: $\quad u_{t+1} \leftarrow (1-\lambda_t)u_t + \lambda_t v_t$
12: **end for**

---

The LCG procedure is a parameter-free algorithm. Note that while line search can be expensive in general, for our subproblems, function evaluation is very cheap. The algorithm needs only one LO oracle call to estimate the initial functional value gap at Line 1. Alternatively, this can be also done approximately via binary search with LOsep. The algorithm maintains a sequence, $\{\Phi_t\}$, that provides valid upper bounds for the functional value gap at the current iterate, i.e., $\phi(u_t) - \phi^* \leq 2\Phi_{t-1}$ (see Theorem 5.1 of Braun et al. (2016)), and it halves the value of $\Phi_t$ only when the current oracle call is negative. Finally, our LCG procedure exits at Line 5 whenever LOsep$_X$ returns a negative call and $\Phi_{t-1} = \eta$, which ensures that $gap_\phi(\bar{u}) = \max_{y \in X}\langle \nabla\phi(\bar{u}), \bar{u} - y \rangle \leq \eta$.

Theorem 1 below provides a bound for the total number of iterations (or calls to the LOsep$_X$ oracle) that the LCG procedure requires to generate a point $\bar{u} \in X$ with $gap_\phi(\bar{u}) \leq \eta$.

**Theorem 1** *Procedure 1 returns a point $\bar{u} \in X$ such that the duality gap at point $\bar{u}$ is bounded by $\eta$, i.e., $gap_\phi(\bar{u}) \leq \eta$. Furthermore, the total number of iterations $T$ (and hence LOsep$_X$ calls) performed by Procedure 1 is at most*

$$T \leq \begin{cases} \kappa + \frac{8\alpha^2 C_\phi}{\eta} + 2, & \eta < \alpha C_\phi; \\ \kappa + 4\alpha + \frac{4\alpha^2 C_\phi}{\eta} + 2, & \eta \geq \alpha C_\phi, \end{cases} \quad (2.12)$$

with $\kappa := 4\alpha \left\lceil \log \frac{\Phi_0}{\alpha C_\phi} \right\rceil + \log \frac{\Phi_0}{\eta}$.

**Proof** From the observations above, it is clear that the duality gap at the output point $\bar{u}$ is bounded by $\eta$.

Also observe that the procedure calls LOsep$_X$ once per iteration. In order to demonstrate the bound in (2.12), we split the LCG procedure into two phases, and bound the number of iterations separately for each phase. Let $C_\phi$ denote the curvature of the smooth convex function $\phi$.

We say Procedure 1 is in the first phase whenever $\Phi_{t-1} > \eta$. In view of Theorem 5.1 in Braun et al. (2016), it is clear that the number of iterations in the first phase can be bounded as

$$T_1 \leq 4\alpha \left\lceil \log \frac{\Phi_0}{\alpha C_\phi} \right\rceil + \frac{4\alpha^2 C_\phi}{\eta} + \log \frac{\Phi_0}{\eta}.$$

Procedure 1 enters the second phase when $\Phi_{t-1} \leq \eta$. Again with the argumentation in Theorem 5.1 in Braun et al. (2016), we obtain that the total number of positive calls in this phase can be bounded by $\frac{4\alpha^2 C_\phi}{\eta}$, if $\eta < \alpha C_\phi$, or by $4\alpha$ if $\eta \geq \alpha C_\phi$. Moreover, the procedure exits whenever the current LOsep$_X$ oracle call is a negative call. Hence, the number of iterations in the second phase can be bounded by

$$T_2 \leq \begin{cases} \frac{4\alpha^2 C_\phi}{\eta} + 1, & \eta < \alpha C_\phi; \\ 4\alpha + 1, & \eta \geq \alpha C_\phi. \end{cases}$$

Thus, our bound in (2.12) can be obtained from the above two bounds plus one more LO oracle call at Line 1. ∎

### 2.3 The convergence properties of CALSGD

This subsection is devoted to establishing the main convergence properties of the CALSGD method. Since the algorithm is stochastic, we will establish the convergence results for finding a stochastic $\epsilon$-solution, i.e., a point $\bar{x} \in X$ s.t. $\mathbb{E}[f(\bar{x}) - f(x^*)] \leq \epsilon$. We first state a simple technical result from Lan and Zhou (2014, Lemma 2.1) that we will use.

**Lemma 2** *Let $w_t \in (0, 1]$, $t = 1, 2, \ldots$, be given. Also let us denote*

$$W_t := \begin{cases} 1 & t = 1 \\ (1 - w_t)W_{t-1} & t \geq 2. \end{cases}$$

Suppose that $W_t > 0$ for all $t \geq 2$ and that the sequence $\{\delta_t\}_{t \geq 0}$ satisfies

$$\delta_t \leq (1 - w_t)\delta_{t-1} + B_t, \quad t = 1, 2, \ldots.$$

Then for any $1 \leq l \leq k$, we have

$$\delta_k \leq W_k \left( \tfrac{1-w_l}{W_l}\delta_{l-1} + \sum_{i=l}^{k} \tfrac{B_i}{W_i} \right).$$

Theorem 3 describes the main convergence properties of the CALSGD method (cf. Algorithm 1).

**Theorem 3** *Let $\Gamma_k$ be defined as follows,*

$$\Gamma_k := \begin{cases} 1 & k = 1 \\ \Gamma_{k-1}(1 - \gamma_k) & k \geq 2. \end{cases} \tag{2.13}$$

*Suppose that $\{\beta_k\}$ and $\{\gamma_k\}$ in the CALSGD algorithm satisfy*

$$\gamma_1 = 1 \quad \text{and} \quad L\gamma_k \leq \beta_k, \quad k \geq 1. \tag{2.14}$$

a) *If*

$$\tfrac{\beta_k \gamma_k}{\Gamma_k} \geq \tfrac{\beta_{k-1} \gamma_{k-1}}{\Gamma_{k-1}}, \quad k \geq 2, \tag{2.15}$$

*then under assumptions (2.2) and (2.3), we have*

$$\mathbb{E}\left[f(y_k) - f(x^*)\right] \leq \tfrac{\beta_k \gamma_k}{2} D_X^2 + \Gamma_k \sum_{i=1}^{k} \left[\tfrac{\eta_i \gamma_i}{\Gamma_i} + \tfrac{\gamma_i \sigma^2}{2\Gamma_i B_i(\beta_i - L\gamma_i)}\right], \tag{2.16}$$

*where $x^*$ is an arbitrary optimal solution of (2.1) and $D_X$ is defined in (1.4).*

b) *If*

$$\tfrac{\beta_k \gamma_k}{\Gamma_k} \leq \tfrac{\beta_{k-1} \gamma_{k-1}}{\Gamma_{k-1}}, \quad k \geq 2, \tag{2.17}$$

*(rather than (2.15)) is satisfied, then the result in part a) holds by replacing $\beta_k \gamma_k D_X^2$ with $\beta_1 \Gamma_k \|x_0 - x^*\|^2$ in the first term of the RHS of (2.16).*

c) *Under the assumptions in part a) or b), the number of inner iterations performed at the k-th outer iterations is bounded by*

$$T_k = \begin{cases} \kappa + \tfrac{8\alpha^2 \beta_k D_X^2}{\eta_k} + 2, & \eta_k < \alpha \beta_k D_X^2; \\ \kappa + 4\alpha + \tfrac{4\alpha^2 \beta_k D_X^2}{\eta_k} + 2, & \eta_k \geq \alpha \beta_k D_X^2, \end{cases} \tag{2.18}$$

*with $\kappa := 4\alpha \left\lceil \log \tfrac{\Phi_0^k}{\alpha \beta_k D_X^2} \right\rceil + \log \tfrac{\Phi_0^k}{\eta_k}.$*

**Proof** Let us denote $\delta_{k,j} = F'(z_k, \xi_{k,j}) - f'(z_k)$ and $\delta_k \equiv g_k - f'(z_k) = \sum_{j=1}^{B_k} \delta_{k,j}/B_k$. We first show part a). In view of (1.6), (2.4) and (2.7), we have

$$f(y_k) \leq l_f(z_k; y_k) + \tfrac{L}{2}\|y_k - z_k\|^2$$
$$= (1-\gamma_k)l_f(z_k; y_{k-1}) + \gamma_k l_f(z_k; x_k) + \tfrac{L\gamma_k^2}{2}\|x_k - x_{k-1}\|^2$$
$$\leq (1-\gamma_k)f(y_{k-1}) + \gamma_k l_f(z_k; x_k) + \tfrac{\beta_k \gamma_k}{2}\|x_k - x_{k-1}\|^2 - \tfrac{\gamma_k(\beta_k - L\gamma_k)}{2}\|x_k - x_{k-1}\|^2,$$

where the last inequality follows from the convexity of $f(\cdot)$. Also observe that by (2.10), we have

$$\langle g_k + \beta_k(x_k - x_{k-1}), x_k - x\rangle \leq \eta_k, \quad \forall x \in X,$$

which implies that

$$\tfrac{1}{2}\|x_k - x_{k-1}\|^2 = \tfrac{1}{2}\|x_{k-1} - x\|^2 - \tfrac{1}{2}\|x_k - x\|^2 - \langle x_{k-1} - x_k, x_k - x\rangle$$
$$\leq \tfrac{1}{2}\|x_{k-1} - x\|^2 - \tfrac{1}{2}\|x_k - x\|^2 + \tfrac{1}{\beta_k}\langle g_k, x - x_k\rangle + \tfrac{\eta_k}{\beta_k}. \tag{2.19}$$

Combing the above two relations, we have

$$f(y_k) \leq (1-\gamma_k)f(y_{k-1}) + \gamma_k l_f(z_k, x_k) + \gamma_k \langle g_k, x - x_k\rangle$$
$$+ \tfrac{\beta_k \gamma_k}{2}\left[\|x_{k-1} - x\|^2 - \|x_k - x\|^2\right]$$
$$+ \eta_k \gamma_k - \tfrac{\gamma_k(\beta_k - L\gamma_k)}{2}\|x_k - x_{k-1}\|^2$$
$$= (1-\gamma_k)f(y_{k-1}) + \gamma_k l_f(z_k, x) + \gamma_k \langle \delta_k, x - x_k\rangle$$
$$+ \tfrac{\beta_k \gamma_k}{2}\left[\|x_{k-1} - x\|^2 - \|x_k - x\|^2\right]$$
$$+ \eta_k \gamma_k - \tfrac{\gamma_k(\beta_k - L\gamma_k)}{2}\|x_k - x_{k-1}\|^2.$$

Using the above inequality and the fact that

$$\langle \delta_k, x - x_k\rangle - \tfrac{(\beta_k - L\gamma_k)}{2}\|x_k - x_{k-1}\|^2 = \langle \delta_k, x - x_{k-1}\rangle + \langle \delta_k, x_{k-1} - x_k\rangle - \tfrac{(\beta_k - L\gamma_k)}{2}\|x_k - x_{k-1}\|^2$$
$$\leq \langle \delta_k, x - x_{k-1}\rangle + \tfrac{\|\delta_k\|_*^2}{2(\beta_k - L\gamma_k)},$$

we obtain for all $x \in X$,

$$f(y_k) \leq (1-\gamma_k)f(y_{k-1}) + \gamma_k f(x) + \eta_k \gamma_k + \tfrac{\beta_k \gamma_k}{2}\left[\|x_{k-1} - x\|^2 - \|x_k - x\|^2\right]$$
$$+ \gamma_k \langle \delta_k, x - x_{k-1}\rangle + \tfrac{\gamma_k \|\delta_k\|_*^2}{2(\beta_k - L\gamma_k)}. \tag{2.20}$$

Subtracting $f(x)$ from both sides of (2.20) and applying Lemma 2, we have

$$f(y_k) - f(x) \leq \Gamma_k(1-\gamma_1)\left[f(y_0) - f(x)\right] + \Gamma_k \sum_{i=1}^{k} \tfrac{\eta_i \gamma_i}{\Gamma_i} + \Gamma_k \sum_{i=1}^{k} \tfrac{\beta_i \gamma_i}{2\Gamma_i}\left[\|x_{k-1} - x\|^2 - \|x_k - x\|^2\right]$$
$$+ \Gamma_k \sum_{i=1}^{k} \tfrac{\gamma_i}{\Gamma_i}\left[\langle \delta_i, x - x_{i-1}\rangle + \tfrac{\|\delta_i\|_*^2}{2(\beta_i - L\gamma_i)}\right]. \tag{2.21}$$

Also observe that

$$\sum_{i=1}^{k} \tfrac{\beta_i \gamma_i}{\Gamma_i}(\|x_{i-1} - x\|^2 - \|x_i - x\|^2)$$
$$= \tfrac{\beta_1 \gamma_1}{\Gamma_1}\|x_0 - x\|^2 - \tfrac{\beta_k \gamma_k}{\Gamma_k}\|x_k - x\|^2 + \sum_{i=2}^{k}\left(\tfrac{\beta_i \gamma_i}{\Gamma_i} - \tfrac{\beta_{i-1}\gamma_{i-1}}{\Gamma_{i-1}}\right)\|x_{i-1} - x\|^2$$
$$\leq \tfrac{\beta_1 \gamma_1}{\Gamma_1}D_X^2 + \sum_{i=2}^{k}\left(\tfrac{\beta_i \gamma_i}{\Gamma_i} - \tfrac{\beta_{i-1}\gamma_{i-1}}{\Gamma_{i-1}}\right)D_X^2$$
$$= \tfrac{\beta_k \gamma_k}{\Gamma_k}D_X^2,$$

where the inequality follows from the third assumption in (2.15) and the definition of $D_X$ in (1.4).

Therefore, from the above two relations and the fact that $\gamma_1 = 1$, we can conclude that

$$f(y_k) - f(x) \leq \tfrac{\beta_k \gamma_k}{2} D_X^2 + \Gamma_k \sum_{i=1}^{k} \tfrac{\gamma_i}{\Gamma_i} \left[ \eta_i + \tfrac{\|\delta_i\|_*^2}{2(\beta_i - L\gamma_i)} + \sum_{j=1}^{B_i} B_i^{-1} \langle \delta_{i,j}, x - x_{i-1} \rangle \right]. \quad (2.22)$$

Note that by our assumptions on SFO, the random variables $\delta_{i,j}$ are independent of the search point $x_{i-1}$ and hence $\mathbb{E}[\langle \delta_{i,j}, x^* - x_{i-1} \rangle] = 0$. In addition, relation (2.8) implies that $\mathbb{E}[\|\delta_i\|_*^2] \leq \sigma^2 / B_i$. Using the previous two observations and taking expectation on both sides of (2.22) (with $x = x^*$) we obtain (2.16).

Similarly, Part b) follows from (2.21), the assumption that $\gamma_1 = 1$, and the fact that

$$\sum_{i=1}^{k} \tfrac{\beta_i \gamma_i}{\Gamma_i} (\|x_{i-1} - x\|^2 - \|x_i - x\|^2) \leq \tfrac{\beta_1 \gamma_1}{\Gamma_1} \|x_0 - x\|^2 - \tfrac{\beta_k \gamma_k}{\Gamma_k} \|x_k - x\|^2 \leq \beta_1 \|x_0 - x\|^2, \quad (2.23)$$

due to the assumptions in (2.14) and (2.17).

Let $\Phi_0^k$ denote the initial bound obtained in Line 1 of the LCG procedure at the $k$-th outer iteration. The result in Part c) follows immediately from (2.12) and the fact that $C_{\psi_k} = \beta_k D_X^2$. ∎

Now we provide two different sets of parameters $\{\beta_k\}, \{\gamma_k\}, \{\eta_k\}$, and $\{B_k\}$, which lead to optimal complexity bounds on the number of calls to the SFO and $\text{LOsep}_X$ oracles.

**Corollary 4** *Suppose that $\{\beta_k\}, \{\gamma_k\}, \{\eta_k\}$, and $\{B_k\}$ in the CALSGD method are set to*

$$\beta_k = \tfrac{4L}{k+2}, \quad \gamma_k = \tfrac{3}{k+2}, \quad \eta_k = \tfrac{LD_X^2}{k(k+1)}, \quad \text{and} \quad B_k = \left\lceil \tfrac{\sigma^2 (k+2)^3}{L^2 D_X^2} \right\rceil, \quad k \geq 1, \quad (2.24)$$

*and we assume $\|f'(x^*)\|$ is bounded for any optimal solution $x^*$ of (2.1). Under assumptions (2.2) and (2.3), we have*

$$\mathbb{E}\left[ f(y_k) - f(x^*) \right] \leq \tfrac{6LD_X^2}{(k+2)^2} + \tfrac{9LD_X^2}{2(k+1)(k+2)}, \quad \forall k \geq 1. \quad (2.25)$$

*As a consequence, the total number of calls to the SFO and $\text{LOsep}_X$ oracles performed by the CALSGD method for finding a stochastic $\epsilon$-solution of (1.1), respectively, can be bounded by*

$$\mathcal{O}\left\{ \sqrt{\tfrac{LD_X^2}{\epsilon}} + \tfrac{\sigma^2 D_X^2}{\epsilon^2} \right\}, \quad (2.26)$$

*and*

$$\mathcal{O}\left\{ \sqrt{\tfrac{LD_X^2}{\epsilon}} \log \tfrac{LD_X^2}{\Lambda \epsilon} + \tfrac{LD_X^2}{\epsilon} \right\} \quad \text{with probability } 1 - \Lambda. \quad (2.27)$$

**Proof** It can be easily seen from (2.24) that (2.14) holds. Also note that by (2.24), we have

$$\Gamma_k = \tfrac{6}{k(k+1)(k+2)}, \quad (2.28)$$

and hence

$$\tfrac{\beta_k \gamma_k}{\Gamma_k} = \tfrac{2Lk(k+1)}{k+2},$$

which implies that (2.15) holds. It can also be easily checked from (2.28) and (2.24) that

$$\sum_{i=1}^{k} \tfrac{\eta_i \gamma_i}{\Gamma_i} \leq \tfrac{kLD_X^2}{2}, \quad \sum_{i=1}^{k} \tfrac{\gamma_i}{\Gamma_i B_i (\beta_i - L\gamma_i)} \leq \tfrac{kLD_X^2}{2\sigma^2}.$$

Using the bound in (2.16), we obtain (2.25), which implies that the total number of outer iterations $N$ can be bounded by $\mathcal{O}\left(\sqrt{LD_X^2/\epsilon}\right)$ under the assumptions (2.2) and (2.3). The bound in (2.26) then immediately follows from this observation and the fact that the number of calls to the SFO oracle is bounded by

$$\sum_{k=1}^{N} B_k \leq \sum_{k=1}^{N} \tfrac{\sigma^2 (k+2)^3}{L^2 D_X^2} + N \leq \tfrac{\sigma^2 (N+3)^4}{4L^2 D_X^2} + N.$$

We now provide a good estimation for $\Phi_0^k$ (cf. Line 1 in LCG procedure) at the $k$-th outer iteration. In view of the definition of $\Phi_0^k$ and $\psi(\cdot)$ (cf. (2.9)), we have,

$$\Phi_0^k = \langle \psi_k'(x_{k-1}), x_{k-1} - x \rangle = \langle g_k, x_{k-1} - x \rangle.$$

Moreover, let $A_k := \|g_k - f'(z_k)\|_* \geq \sqrt{\tfrac{N\sigma^2}{\Lambda B_k}}$, by Chebyshev's inequality and (2.8), we obtain,

$$\mathrm{Prob}\{A_k\} \leq \tfrac{\mathbb{E}[\|g_k - f'(z_k)\|_*^2] \Lambda B_k}{N\sigma^2} \leq \tfrac{\Lambda}{N}, \ \forall \Lambda < 1, k \geq 1,$$

which implies that $\mathrm{Prob}\{\bigcap_{k=1}^{N} \bar{A}_k\} \geq 1 - \Lambda$. Hence, by Cauchy-Schwarz and triangle inequalities, we have with probability $1 - \Lambda$,

$$\begin{aligned}
\Phi_0^k &= \langle g_k - f'(z_k), x_{k-1} - x \rangle + \langle f'(z_k), x_{k-1} - x \rangle\} \\
&\leq \left( \sqrt{\tfrac{N\sigma^2}{\Lambda B_k}} + \|f'(z_k) - f'(x^*)\|_* + \|f'(x^*)\|_* \right) D_X \\
&\leq \left( \sqrt{\tfrac{N}{\Lambda k^3}} + 1 \right) LD_X^2 + \|f'(x^*)\|_* D_X, \quad (2.29)
\end{aligned}$$

where the last inequality follows from (1.6) and (2.24).

Note that we always have $\eta_k < \alpha \beta_k D_X^2$. Therefore, it follows from the bound in (2.18), (2.24), and (2.29) that the total number of inner iterations can be bounded by

$$\begin{aligned}
\sum_{k=1}^{N} T_k &\leq \sum_{k=1}^{N} \left[ 4\alpha \left( \log \tfrac{\Phi_0^k}{\alpha \beta_k D_X^2} + 1 \right) + \log \tfrac{\Phi_0^k}{\eta_k} + \tfrac{8\alpha^2 \beta_k D_X^2}{\eta_k} + 2 \right] \\
&\leq \sum_{k=1}^{N} \left[ 5\alpha \log \left( 2k^2 \left( \sqrt{\tfrac{N}{\Lambda k^3}} + 1 + \tfrac{\|f'(x^*)\|_*}{LD_X} \right) \right) + 32\alpha^2 k \right] + (4\alpha + 2) N \\
&= \mathcal{O}\left( N \log \tfrac{N^2}{\Lambda} + N^2 + N \right),
\end{aligned}$$

which implies that our bound in (2.27). ∎

We now provide a slightly improved complexity bound on the number of calls to the SFO oracle which depends on the distance from the initial point to the set of optimal solutions, rather than the diameter $D_X$. In order to obtain this improvement, we need to estimate $D_0 \geq \|x_0 - x^*\|$ and to fix the number of iterations $N$ in advance. This result will play an important role for the analysis of the CALSGD method to solve strongly convex problems (see Section 4.1).

**Corollary 5** *Suppose that there exists an estimate $D_0$ s.t. $\|x_0 - x^*\| \leq D_0 \leq D_X$. Also assume that the outer iteration limit $N \geq 1$ is given. If*

$$\beta_k = \tfrac{3L}{k}, \quad \gamma_k = \tfrac{2}{k+1}, \quad \eta_k = \tfrac{2LD_0^2}{Nk}, \quad and \quad B_k = \left\lceil \tfrac{\sigma^2 N(k+1)^2}{L^2 D_0^2} \right\rceil, \quad k \geq 1. \tag{2.30}$$

*Under assumptions (2.2) and (2.3),*

$$\mathbb{E}\left[f(y_N) - f(x^*)\right] \leq \tfrac{8LD_0^2}{N(N+1)}, \quad \forall N \geq 1.$$

*As a consequence, the total number of calls to the* SFO *and* $\mathrm{LOsep}_X$ *oracles performed by the CALSGD method for finding a stochastic $\epsilon$-solution of (1.1), respectively, can be bounded by*

$$\mathcal{O}\left\{\sqrt{\tfrac{LD_0^2}{\epsilon}} + \tfrac{\sigma^2 D_0^2}{\epsilon^2}\right\}, \tag{2.31}$$

*and (2.27).*

**Proof** The proof is similar to Corollary 4, and hence details are skipped. ∎

It should be pointed out that the complexity bound for the number of calls to the LOsep oracle in (2.27) is established with probability $1 - \Lambda$. However, the probability parameter $\Lambda$ only appears in the non-dominant term.

## 3. Deterministic CALSGD

Our goal in this section is to present a deterministic version of CALSGD, which we refer to as CALGD. Instead of calling the SFO oracle to compute the stochastic gradients, we assume that we have access to the exact gradients of $f$. Therefore, the CALGD method calls the FO oracle to obtain the exact gradients $f'(z_k)$ at the $k$-th outer iteration.

The CALGD method is formally described as follows.

---
**Algorithm 2** The conditional accelerated lazy gradient descent (CALGD) method
---
This algorithm is the same as Algorithm 1 except that steps (2.5) and (2.6) are replaced by
$$x_k = \mathrm{LCG}(f'(z_k), \beta_k, x_{k-1}, \alpha, \eta_k). \tag{3.1}$$
---

Similarly to the stochastic case, we can easily see that $x_k$ obtained in (3.1) is an approximate solution for the gradient sliding subproblem

$$\min_{x \in X} \left\{\psi_k(x) := \langle f'(z_k), x \rangle + \tfrac{\beta_k}{2}\|x - x_{k-1}\|^2\right\} \tag{3.2}$$

such that for all $x \in X$

$$\langle \psi_k'(x_k), x_k - x \rangle = \langle f'(z_k) + \beta_k(x_k - x_{k-1}), x_k - x \rangle \leq \eta_k, \tag{3.3}$$

for some $\eta_k \geq 0$.

Theorem 6 describes the main convergence properties of the above CALGD method.

LAN, POKUTTA, ZHOU AND ZINK**Theorem 6** *Let $\Gamma_k$ be defined as in (2.13). Suppose that $\{\beta_k\}$ and $\{\gamma_k\}$ in the CALGD algorithm satisfy (2.14).*

   a) *If (2.15) is satisfied, then for $k \geq 1$,*

$$f(y_k) - f(x^*) \leq \tfrac{\beta_k \gamma_k}{2} D_X^2 + \Gamma_k \sum_{i=1}^{k} \tfrac{\eta_i \gamma_i}{\Gamma_i}. \tag{3.4}$$

   *where $x^*$ is an arbitrary optimal solution of (1.1) and $D_X$ is defined in (1.4).*

   b) *If (2.17) (rather than (2.15)) is satisfied, then for $k \geq 1$,*

$$f(y_k) - f(x^*) \leq \tfrac{\beta_1 \Gamma_k}{2} \|x_0 - x^*\|^2 + \Gamma_k \sum_{i=1}^{k} \tfrac{\eta_i \gamma_i}{\Gamma_i}. \tag{3.5}$$

   c) *Under the assumptions in either part a) or b), the number of inner iterations performed at the $k$-th outer iteration can be bounded by (2.18).*

**Proof** Since the convergence results stated in Theorem 3 cover the deterministic case when we set $\delta_{k,j} = F'(z_k, \xi_{k,j}) - f'(z_k) \equiv 0$, Part a) immediately follows from (2.16) with $\sigma = 0$. Similarly, Part b) follows from (2.21), (2.23) and $\delta_i = \sum_{j=1}^{B_i} \delta_{i,j} = 0$. The proof of Part c) is exactly the same as that of Theorem 3.c). ■

Clearly, there exist various options to specify the parameters $\{\beta_k\}$, $\{\gamma_k\}$, and $\{\eta_k\}$ so as to guarantee the convergence of the CALGD method. In the following corollaries, we provide two different parameter settings for $\{\beta_k\}$, $\{\gamma_k\}$, and $\{\eta_k\}$, which lead to optimal complexity bounds on the total number of calls to the FO and LOsep oracles for smooth convex optimization.

**Corollary 7** *If $\{\beta_k\}$, $\{\gamma_k\}$, and $\{\eta_k\}$ in the CALGD method are set to*

$$\beta_k = \tfrac{3L}{k+1}, \quad \gamma_k = \tfrac{3}{k+2}, \quad \text{and} \quad \eta_k = \tfrac{LD_X^2}{k(k+1)}, \quad \forall k \geq 1, \tag{3.6}$$

*and we assume that $\|f'(x^*)\|$ is bounded for any optimal solution $x^*$ of (1.1), then for any $k \geq 1$,*

$$f(y_k) - f(x^*) \leq \tfrac{15 L D_X^2}{2(k+1)(k+2)}. \tag{3.7}$$

*As a consequence, the total number of calls to the FO and LOsep oracles performed by the CALGD method for finding an $\epsilon$-solution of (1.1) can be bounded by $\mathcal{O}\left(\sqrt{L D_X^2 / \epsilon}\right)$ and $\mathcal{O}\left(L D_X^2 / \epsilon\right)$ respectively.*

**Proof** It can be easily seen from (3.6) that (2.14) holds, $\Gamma_k$ is given by (2.28), and

$$\tfrac{\beta_k \gamma_k}{\Gamma_k} = \tfrac{9L}{(k+1)(k+2)} \tfrac{k(k+1)(k+2)}{6} = \tfrac{3Lk}{2},$$

which implies that (2.15) is satisfied. It then follows from Theorem 6.a), (3.6), and (2.28) that

$$\begin{aligned} f(y_k) - f(x^*) &\leq \tfrac{9 L D_X^2}{2(k+1)(k+2)} + \tfrac{6}{k(k+1)(k+2)} \sum_{i=1}^{k} \tfrac{\eta_i \gamma_i}{\Gamma_i} \\ &= \tfrac{15 L D_X^2}{2(k+1)(k+2)}, \end{aligned}$$

which implies that the total number of outer iterations performed by the CALGD method for finding an $\epsilon$-solution can be bounded by $N = \sqrt{15LD_X^2/(2\epsilon)}$.

We first provide a valid upper bound for $\Phi_0^k$ defined in Line 1 when the CALGD method enters the LCG procedure at the $k$-th outer iteration. In view of the definitions of $\Phi_0^k$ and $\psi(\cdot)$ at Line 1 and (3.2), respectively, we have, for any $k \geq 1$,

$$\begin{aligned}\Phi_0^k &= \langle \psi_k'(x_{k-1}), x_{k-1} - x \rangle = \langle f'(z_k), x_{k-1} - x \rangle \\ &\leq (\|f'(z_k) - f'(x^*)\| + \|f'(x^*)\|)\|x_{k-1} - x\| \\ &\leq LD_X^2 + \|f'(x^*)\|D_X, \end{aligned} \quad (3.8)$$

where the first inequality follows from Cauchy-Schwarz and the triangle inequality, and the second inequality follows from (1.2) and (1.4). Note that we always have $\eta_k < \alpha\beta_k D_X^2$. Therefore, similar to the stochastic case, our $\mathcal{O}(LD_X^2/\epsilon)$ bound immediately follows from the above relation, (2.18), and (3.6). ∎

As before in the stochastic case, we can slightly improve the complexity bound on the calls to the FO oracle in terms of the dependence on $D_X$.

**Corollary 8** *Suppose that there exists an estimate $D_0 \geq \|x_0 - x^*\|$ and that the outer iteration limit $N \geq 1$ is given. If*

$$\beta_k = \tfrac{2L}{k}, \quad \gamma_k = \tfrac{2}{k+1}, \quad \eta_k = \tfrac{2LD_0^2}{Nk}, \quad (3.9)$$

*for $k \geq 1$, then*

$$f(y_N) - f(x^*) \leq \tfrac{6LD_0^2}{N(N+1)}. \quad (3.10)$$

*As a consequence, the total number of calls to the FO and LOsep oracles performed by the CALGD method for finding an $\epsilon$-solution of (1.1) can be bound by*

$$\mathcal{O}\left(D_0\sqrt{\tfrac{L}{\epsilon}}\right) \quad \text{and} \quad \mathcal{O}\left(\tfrac{LD_X^2}{\epsilon}\right) \quad (3.11)$$

*respectively.*

**Proof** The proof is similar to Corollary 7, and hence omitted. ∎

## 4. Generalizations to other optimization problems

We generalize the CALGD and CALSGD methods to solve two other classes of problems frequently seen in machine learning. In particular, we discuss the CALGD method with a restarting technique for solving smooth and strongly convex problems in Subsection 4.1, and in Subsection 4.3 we extend the CALGD method to solve a special class of non-smooth problems. Discussions for the similar extensions for CALSGD method can be found in Subsection 4.2 and 4.4. In addition to the special class of non-smooth problems, we also consider the general non-smooth problems in Subsection 4.5.

### 4.1 Strongly convex optimization

In this subsection, we assume that the objective function $f$ is not only smooth (i.e., (1.6) holds), but also strongly convex, that is, $\exists\ \mu > 0$ s.t.

$$f(y) - f(x) - \langle f'(x), y - x \rangle \geq \tfrac{\mu}{2}\|y - x\|^2, \quad \forall x, y \in X. \tag{4.1}$$

For simplicity, we first establish the convergence results for the deterministic case, i.e., we have access to the exact gradients of the objective function $f$.

The shrinking conditional gradient method in Lan (2013) needs to make additional assumptions on the LO oracle to obtain a linear rate of convergence. However, we will show now that CALGD (relying on the vanilla weak separation oracle) can obtain a linear rate of convergence in terms of the number of calls to the FO oracle and $\mathcal{O}(LD_X^2/\epsilon)$ rate of convergence in the total number of calls to the LOsep oracle. In view of the lower complexity bound established for the LO oracle to solve strongly convex problems in Jaggi (2013) and Lan (2013), our bound for the LOsep oracle is not improvable.

We are now ready to formally describe the CALGD method for solving strongly convex problems, which is obtained by properly restarting the CALGD method (Algorithm 2).

---

**Algorithm 3** The CALGD method for strongly convex problems
---

**Input:** Initial point $p_0 \in X$ and an estimate $\delta_0 > 0$ satisfying $f(p_0) - f(x^*) \leq \delta_0$.
**for** $s = 1, 2, \ldots$ **do**
  Call the CALGD method in Algorithm 2 with input

$$x_0 = p_{s-1} \quad \text{and} \quad N = \left\lceil 2\sqrt{\tfrac{6L}{\mu}} \right\rceil, \tag{4.2}$$

  and parameters

$$\beta_k = \tfrac{2L}{k}, \quad \gamma_k = \tfrac{2}{k+1}, \quad \text{and} \quad \eta_k = \eta_{s,k} := \tfrac{8L\delta_0 2^{-s}}{\mu N k}, \tag{4.3}$$

  and let $p_s$ be its output solution.
**end for**

---

In Algorithm 3, we restart the CALGD method for smooth optimization (i.e., Algorithm 2) every $\lceil 2\sqrt{6L/\mu} \rceil$ iterations. We call each loop iteration a phase of the above CALGD algorithm. Observe that $\{\eta_k\}$ decrease by a factor of 2 as $s$ increments by 1, while $\{\beta_k\}$ and $\{\gamma_k\}$ remain the same. The following theorem shows the convergence of the above variant of the CALGD method.

**Theorem 9** *Assume (4.1) holds and let $\{p_s\}$ be generated by Algorithm 3. Then,*

$$f(p_s) - f(x^*) \leq \delta_0 2^{-s}, \quad s \geq 0.$$

*As a consequence, the total number of calls to the FO and LOsep oracles performed by this algorithm for finding an $\epsilon$-solution of problem (1.1) can be bounded by*

$$\mathcal{O}\left\{ \sqrt{\tfrac{L}{\mu}} \left\lceil \log_2 \max\left(1, \tfrac{\delta_0}{\epsilon}\right) \right\rceil \right\} \quad \text{and} \quad \mathcal{O}\left\{ \tfrac{LD_X^2}{\epsilon} \right\}, \tag{4.4}$$

*respectively.*

**Proof** Denote the total number of phases performed by CALGD method to obtain an $\epsilon$-solution of (1.1) by $S$. In view of the complexity results obtained in Theorem 2.5 in Lan and Zhou (2014), we conclude that

$$S = \left\lceil \log_2 \max\left(1, \tfrac{\delta_0}{\epsilon}\right) \right\rceil. \tag{4.5}$$

The total number of calls to the FO oracle performed by Algorithm 3 is clearly bounded by $NS$, which immediately implies our first result in (4.4).

Now, let $T_{s,k}$ denote the number of calls to the LOsep oracle required at the $k$-th outer iteration in the $s$-th phase. It follows from Theorem 6.c), (3.8), and (4.3) that

$$T_{s,k} \leq \mathcal{O}\left(\tfrac{\beta_k D_X^2}{\eta_{s,k}}\right) = \mathcal{O}\left(\tfrac{\mu D_X^2 2^s N}{\delta_0}\right).$$

Therefore, the total number of calls to the LOsep oracle can be bounded by

$$\begin{aligned}
\sum_{s=1}^{S}\sum_{k=1}^{N} T_{s,k} &\leq \sum_{s=1}^{S}\sum_{k=1}^{N} \mathcal{O}\left(\tfrac{\mu D_X^2 2^s N}{\delta_0}\right) \\
&= \mathcal{O}\left(\tfrac{\mu D_X^2 N^2}{\delta_0}\sum_{s=1}^{S} 2^s\right) \\
&= \mathcal{O}\left(\tfrac{\mu D_X^2 N^2}{\delta_0} 2^{S+1}\right) \\
&= \mathcal{O}\left(\tfrac{\mu D_X^2 N^2}{\epsilon}\right),
\end{aligned}$$

which implies our second bound in (4.4) due to the definitions of $N$ and $S$ in (4.2) and (4.5), respectively. ∎

In view of classic complexity theory for convex optimization, the bound on the total number of calls to the FO oracle (cf. first bound in (4.4)) is optimal for strongly convex optimization. Moreover, in view of the complexity results established in Lan (2013) and the fact that the LOsep oracle is weaker than the LO oracle, the bound on the total number of calls to the LOsep oracle (cf. second bound in (4.4)) is not improvable either.

### 4.2 Strongly convex stochastic optimization

Similarly to the deterministic case we present an optimal algorithm for solving stochastic smooth and strongly convex problems.

---

**Algorithm 4** The CALSGD method for solving strongly convex problems

**Input:** Initial point $p_0 \in X$ and an estimate $\delta_0 > 0$ satisfying $f(p_0) - f(x^*) \le \delta_0$.

**for** $s = 1, 2, \ldots$ **do**

  Call the CALSGD method in Algorithm 1 with input

$$x_0 = p_{s-1} \quad \text{and} \quad N = \left\lceil 4\sqrt{\tfrac{2L}{\mu}}\, \right\rceil, \tag{4.6}$$

  and parameters

$$\beta_k = \tfrac{3L}{k}, \ \gamma_k = \tfrac{2}{k+1}, \ \eta_k = \eta_{s,k} := \tfrac{8L\delta_0 2^{-s}}{\mu N k},$$
$$\text{and } B_k = B_{s,k} := \left\lceil \tfrac{\mu\sigma^2 N (k+1)^2}{4L^2 \delta_0 2^{-s}} \right\rceil, \tag{4.7}$$

  and let $p_s$ be its output solution.

**end for**

---

The main convergence properties of Algorithm 4 are as follows.

**Theorem 10** *Assume that* (4.1) *holds and let* $\{p_s\}$ *be generated by Algorithm 4. Then,*

$$\mathbb{E}[f(p_s) - f(x^*)] \le \delta_0 2^{-s}, \quad s \ge 0.$$

*As a consequence, the total number of calls to the* SFO *and* LOsep *oracles performed by this algorithm for finding a stochastic $\epsilon$-solution of problem* (1.1)-(2.1) *can be bounded by*

$$\mathcal{O}\left\{ \tfrac{\sigma^2}{\mu\epsilon} + \sqrt{\tfrac{L}{\mu}} \left\lceil \log_2 \max\left(1, \tfrac{\delta_0}{\epsilon}\right) \right\rceil \right\}, \tag{4.8}$$

*and*

$$\mathcal{O}\left\{ \tfrac{LD_X^2}{\epsilon} \right\}, \ \text{with probability } 1 - \Lambda, \tag{4.9}$$

*respectively.*

**Proof** In view of Corollary 5, and Theorem 3.4 in Lan and Zhou (2014), the total number of phases, $S$, performed by CALSGD method to find a stochastic $\epsilon$-solution of problem (1.1)-(2.1) is bounded by (4.5). Since the number of outer iterations in each phase is at most $N$, the total number of calls to the SFO oracle is bounded by

$$\sum_{s=1}^{S}\sum_{k=1}^{N} B_k \le \sum_{s=1}^{S}\sum_{k=1}^{N} \left( \tfrac{\mu\sigma^2 N(k+1)^2}{4L^2 \delta_0 2^{-s}} + 1 \right)$$
$$\le \tfrac{\mu\sigma^2 N(N+1)^3}{12 L^2 \delta_0} \sum_{s=1}^{S} 2^s + SN$$
$$\le \tfrac{\mu\sigma^2 N(N+1)^3}{3 L^2 \epsilon} + SN.$$

Moreover, similar to (2.29), we obtain a good estimator for $\Phi_0^{s,k}$, for any $0 < \Lambda \le 1$

$$\Phi_0^{s,k} \le \left( \sqrt{\tfrac{4SL^2\delta_0}{\Lambda \mu k^2 2^s}} + 1 \right) LD_X^2 + \|f'(x^*)\|_* D_X,$$

with probability $1-\Lambda$. Let $T_{s,k}$ denote the number of calls to the LOsep oracle required at the $k$-th outer iteration in the $s$-th phase of the CALSGD method. It follows from Theorem 3.c), the above relation, and (4.7) that with probability $1 - \Lambda$,

$$T_{s,k} \leq \mathcal{O}\left(\log \tfrac{\Phi_0^{s,k}}{\eta_{s,k}} + \tfrac{\beta_k D_X^2}{\eta_{s,k}}\right) = \mathcal{O}\left(\tfrac{\mu D_X^2 2^s N}{\delta_0}\right)$$

holds. Therefore, the total number of calls to the LOsep oracle is bounded by

$$\begin{aligned}\sum_{s=1}^{S}\sum_{k=1}^{N}T_{s,k} &\leq \sum_{s=1}^{S}\sum_{k=1}^{N}\mathcal{O}\left(\tfrac{\mu D_X^2 2^s N}{\delta_0}\right)\\ &= \mathcal{O}\left(\mu D_X^2 N^2 \delta_0^{-1}\sum_{s=1}^{S}2^s\right)\\ &= \mathcal{O}\left(\tfrac{\mu D_X^2 N^2}{\epsilon}\right),\end{aligned}$$

which implies the bound in (4.9), due to the definitions of $N$ and $S$ in (4.6) and (4.5), respectively. ∎

According to Theorem 10, the total number of calls to the SFO oracle is bounded by $\mathcal{O}(1/\epsilon)$, which is optimal in view of the classic complexity theory for strongly convex optimization (see (Ghadimi and Lan, 2012, 2013)). Moreover, the total number of calls to the LOsep oracle is bounded by $\mathcal{O}(1/\epsilon)$, which is the same bound as for the CALGD method for strongly convex optimization and hence not improvable.

### 4.3 Non-smooth optimization: Saddle point problems

For the sake of simplicity, we consider the deterministic case, i.e., the problem of interest is an important class of saddle point problems with $f$ given in the form of

$$f(x) = \max_{y \in Y}\left\{\langle Ax, y\rangle - \hat{f}(y)\right\}, \tag{4.10}$$

where $A : \mathbb{R}^n \to \mathbb{R}^m$ denotes a linear operator, $Y \in \mathbb{R}^m$ is a convex compact set, and $\hat{f} : Y \to \mathbb{R}$ is a simple convex function. Since the objective function $f$ is non-smooth, we cannot directly apply the CALGD method presented in the previous section. However, as shown by Nesterov (2005), the function $f(\cdot)$ in (4.10) can be closely approximated by a class of smooth convex functions. More specifically, let $\omega : Y \to \mathbb{R}$ be a given strongly convex function with strongly convex modulus $\sigma_\omega > 0$, i.e.,

$$\omega(y) \geq \omega(x) + \langle \omega'(x), y - x\rangle + \tfrac{\sigma_\omega}{2}\|y - x\|^2, \forall x, y \in Y,$$

and let us denote $c_\omega := \operatorname{argmin}_{y \in Y}\omega(y)$, $W(y) := \omega(y) - \omega(c_\omega) - \langle \nabla\omega(c_\omega), y - c_\omega\rangle$ and

$$\mathcal{D}_{Y,W}^2 := \max_{y \in Y} W(y).$$

It can be easily seen that

$$\|y - c_\omega\|^2 \leq \tfrac{2}{\sigma_\omega}W(y) \leq \tfrac{2}{\sigma_\omega}\mathcal{D}_{Y,W}^2, \quad \forall y \in Y,$$

and hence that
$$\|y_1 - y_2\|^2 \leq \tfrac{4}{\sigma_\omega}\mathcal{D}^2_{Y,W}, \quad \forall y_1, y_2 \in Y.$$

In view of these relations, the function $f(\cdot)$ in (4.10) can be closely approximated by
$$f_\tau(x) := \max_{y \in Y}\left\{\langle Ax, y\rangle - \hat{f}(y) - \tau\left[W(y) - \mathcal{D}^2_{Y,W}\right]\right\}. \tag{4.11}$$

In particular, for any $\tau \geq 0$,
$$f(x) \leq f_\tau(x) \leq f(x) + \tau\mathcal{D}^2_{Y,W}, \quad \forall x \in X.$$

Moreover, Nesterov (2005) shows that $f_\tau(\cdot)$ is differentiable and its gradients are Lipschitz continuous with the Lipschitz constant given by
$$\mathcal{L}_\tau := \tfrac{\|A\|^2}{\tau\sigma_\omega}. \tag{4.12}$$

Throughout this subsection, we assume that the feasible region $Y$ and the function $\hat{f}$ are simple enough, so that the subproblem in (4.11) is easy to solve. Therefore, the major computational cost for gradient calculations of $f_\tau$ lie in the evaluations of the linear operator $A$ and its adjoint operator $A^T$. We are now ready to present a variant of the CALGD method, which can achieve optimal bounds on the number of calls to the LOsep oracle and the number of evaluations of the linear operators $A$ and $A^T$.

---

**Algorithm 5** The CALGD method for solving saddle point problems

This algorithm is the same as Algorithm 2 except that (3.1) is replaced by
$$x_k = \text{LCG}(f'_{\tau_k}(z_k), \beta_k, x_{k-1}, \alpha, \eta_k), \tag{4.13}$$
for some $\tau_k \geq 0$.

---

In Theorem 11 we state the main convergence properties of this modified CALGD method to solve the saddle point problem in (1.1)-(4.10).

**Theorem 11** *Suppose that $\tau_1 \geq \tau_2 \geq \ldots \geq 0$. Also assume that $\{\beta_k\}$ and $\{\gamma_k\}$ satisfy (2.14) (with $L$ replaced by $L_{\tau_k}$ defined in (4.12)) and (2.15). Then, for all $k \geq 1$,*
$$f(y_k) - f(x^*) \leq \tfrac{\beta_k\gamma_k}{2}D^2_X + \Gamma_k\sum_{i=1}^k \tfrac{\gamma_i}{\Gamma_i}\left(\eta_i + \tau_i\mathcal{D}^2_{Y,W}\right), \tag{4.14}$$

*where $x^*$ is an arbitrary optimal solution of (1.1)-(4.10). Moreover, the number of inner iterations performed at the $k$-th outer iteration is bounded by (2.18).*

**Proof** The proof is similar to Theorem 4.1 in Lan and Zhou (2014), and hence omitted. ∎

We now provide two different sets of parameter settings for $\{\beta_k\}, \{\gamma_k\}, \{\eta_k\}$, and $\{\tau_k\}$ which can guarantee the optimal convergence of the above variant of the CALGD method for saddle point optimization. Specifically, Corollary 12 gives a static setting for parameter $\{\tau_k\}$ under the assumption that the outer iteration limit $N \geq 1$ is given, while a dynamic setting is provided in Corollary 13.

**Corollary 12** *Assume the outer iteration limit $N \geq 1$ is given. If*

$$\tau_k \equiv \tau = \frac{2\|A\|D_X}{\mathcal{D}_{Y,W}\sqrt{\sigma_\omega}N}, \quad k \geq 1, \tag{4.15}$$

*and $\{\beta_k\}$, $\{\gamma_k\}$, and $\{\eta_k\}$ used in Algorithm 5 are set to*

$$\beta_k = \frac{3\mathcal{L}_{\tau_k}}{k+1}, \quad \gamma_k = \frac{3}{k+2}, \quad \text{and} \quad \eta_k = \frac{\mathcal{L}_{\tau_k}D_X^2}{k^2}, \quad k \geq 1, \tag{4.16}$$

*then the number of linear operator evaluations (for $A$ and $A^T$) and the number of calls to the LOsep oracle performed by Algorithm 5 for finding an $\epsilon$-solution of problem (1.1)-(4.10), respectively, is bounded by*

$$\mathcal{O}\left\{\frac{\|A\|D_X\mathcal{D}_{Y,W}}{\sqrt{\sigma_\omega}\epsilon}\right\} \quad \text{and} \quad \mathcal{O}\left\{\frac{\|A\|^2D_X^2\mathcal{D}_{Y,W}^2}{\sigma_\omega\epsilon^2}\right\}. \tag{4.17}$$

**Proof** In view of the result in Corollary 4.2 of Lan and Zhou (2014), our first bound in (4.17) immediately follows. Moreover, it follows from (3.8), (2.18), (4.15), (4.16) and (4.12) that the total number of calls to the LOsep oracle is bounded by

$$\sum_{k=1}^{N}T_k \leq \sum_{k=1}^{N}\mathcal{O}\left(\frac{\beta_k D_X^2}{\eta_k}\right) = \sum_{k=1}^{N}\mathcal{O}\left(\frac{\mathcal{L}_{\tau_k}D_X^2}{k+1}\frac{k^2}{\mathcal{L}_{\tau_k}D_X^2}\right) = \mathcal{O}(N^2),$$

which implies our second bound in (4.17). ∎

**Corollary 13** *Suppose that parameter $\{\tau_k\}$ is now set to*

$$\tau_k = \frac{2\|A\|D_X}{\mathcal{D}_{Y,W}\sqrt{\sigma_\omega}k}, \quad k \geq 1, \tag{4.18}$$

*and the parameters $\{\beta_k\}$, $\{\gamma_k\}$, and $\{\eta_k\}$ used in Algorithm 5 are set as in (4.16). Then, the number of linear operator evaluations (for $A$ and $A^T$) and the number of calls to the LOsep oracle performed by Algorithm 5 for finding an $\epsilon$-solution of problem (1.1)-(4.10) is bounded by the two bounds as given in (4.17) respectively.*

**Proof** The proof is similar to the Corollary 12, and hence omitted. ∎

In view of the discussions in Chen et al. (2014), the obtained bound on the total number of operator evaluations (cf. first bound in (4.17)) is not improvable for solving the saddle point problems in (1.1)-(4.10). Moreover, according to Lan (2013) and the fact that the LOsep oracle is weaker than LO oracle, the $\mathcal{O}(1/\epsilon^2)$ bound on the total number of calls to the LOsep is not improvable.

### 4.4 Non-smooth stochastic optimization: stochastic saddle point problems

In this subsection, we briefly discuss stochastic saddle point problems, i.e., only stochastic gradients of $f_\tau$ (cf. (4.11)) are available. In particular, we consider the situation when the original objective function $f$ in (1.1) is given by

$$f(x) = \mathbb{E}\left[\max_{y\in Y}\langle A_\xi x, y\rangle - \hat{f}(y,\xi)\right], \tag{4.19}$$

where $\hat{f}(\cdot, \xi)$ is simple concave function for all $\xi \in \Xi$ and $A_\xi$ is a random linear operator such that
$$\mathbb{E}\left[\|A_\xi\|^2\right] \leq L_A^2 \tag{4.20}$$
We can solve this stochastic saddle point problem by replacing (4.13) with
$$x_k = \text{LCG}(g_k, x_{k-1}, \beta_k, \eta_k),$$
where $g_k = \frac{1}{B_k} \sum_{j=1}^{B_k} F'_{\tau_k}(z_k, \xi_{k,j})$ for some $\tau_k \geq 0$ and $B_k \geq 1$. By properly specifying $\{\beta_k\}$, $\{\eta_k\}$, $\{\tau_k\}$, and $\{B_k\}$, we can show that the number of linear operator evaluations (for $A_\xi$ and $A_\xi^T$) and the number of calls to the LOsep oracle performed by this variant of CALSGD method for finding a stochastic $\epsilon$-solution of problem (1.1)-(4.19) is bounded by
$$\mathcal{O}\left\{\frac{L_A^2 D_X^2 \mathcal{D}_{Y,W}^2}{\sigma_\omega \epsilon^2}\right\},$$
and
$$\mathcal{O}\left\{\frac{L_A^2 D_X^2 \mathcal{D}_{Y,W}^2}{\sigma_\omega \epsilon^2}\right\}$$
with probability $1 - \Lambda$ respectively. This result can be proved by combining the techniques in Section 2 and those in Theorem 11. However, we skip the details of these developments for the sake of simplicity.

### 4.5 General non-smooth stochastic optimization

In this subsection, we present a variant of CALSGD for solving general non-smooth stochastic problems. Observe that the online Frank-Wolfe method proposed in Hazan and Kale (2012) needs $\mathcal{O}(1/\epsilon^4)$ number of calls to both the SFO and LO oracles, however, this variant of CALSGD improves the bound for SFO oracle to an optimal bound of $\mathcal{O}(1/\epsilon^2)$ while still maintaining a comparable bound $\mathcal{O}(1/\epsilon^4)$ on the number of calls to $\text{LOsep}_X$ for solving general non-smooth stochastic problems. Throughout this subsection, we assume that the objective function $f$ is Lipschitz continuous, i.e., $\exists M > 0$ s.t.
$$f(y) \leq f(x) + \langle f'(x), y - x \rangle + M\|x - y\|, \; \forall x, y \in X, \tag{4.21}$$
for $f'(x) \in \partial f(x)$, where $\partial f(x)$ denotes the sub-differential of $f$ at $x$. We also assume that the SFO oracle, for a given search point $z_k \in X$, outputs a stochastic sub-gradient $F'(z_k, \xi_k)$ such that (2.2) and (2.3) hold.

Algorithm 6 below is a variant of the CALSGD method, which can achieve optimal complexity bounds on the number of calls to both the $\text{LOsep}_X$ and SFO oracles.

---
**Algorithm 6** The CALSGD method for solving general non-smooth stochastic problems

This algorithm is the same as Algorithm 1 except that $B_k \equiv 1$ and $F'(z_k, \xi_{k,j})$ in (2.5) is replaced by a stochastic sub-gradient computed by the SFO at $z_k$.

---

The following theorem establishes the convergence of the above variant of the CALSGD method.

CONDITIONAL ACCELERATED LAZY STOCHASTIC GRADIENT DESCENT

**Theorem 14** *Assume (2.2), (2.3) and (4.21) hold, and let $\Gamma_k$ be defined as in (2.13) and $\gamma_1 = 1$.*

*a) If (2.15) is satisfied, then for $k \geq 1$, we have*

$$\mathbb{E}\left[f(y_k) - f(x^*)\right] \leq \frac{\beta_k \gamma_k}{2} D_X^2 + \Gamma_k \sum_{i=1}^{k} \left[\frac{\eta_i \gamma_i}{\Gamma_i} + \frac{\gamma_i(\sigma^2 + M^2)}{\Gamma_i \beta_i}\right], \quad (4.22)$$

*where $x^*$ is an arbitrary optimal solution of (2.1) and $D_X$ is defined in (1.4).*

*b) If (2.17) (rather than (2.15)) is satisfied, then the result in part a) holds by replacing $\beta_k \gamma_k D_X^2$ with $\beta_1 \Gamma_k \|x_0 - x^*\|^2$ in the first term of the RHS of (4.22).*

*c) Under the assumptions in part a) or b), the number of inner iterations performed at the k-th outer iterations is bounded by (2.18)*

**Proof** Similar to the proof of Theorem 3, let us denote $\delta_k \equiv g_k - f'(z_k)$. We first show part a). In view of (4.21), (2.4) and (2.7), we have

$$f(y_k) \leq l_f(z_k; y_k) + M\|y_k - z_k\|$$
$$\leq (1 - \gamma_k) f(y_{k-1}) + \gamma_k l_f(z_k; x_k) + M\gamma_k \|x_k - x_{k-1}\|.$$

Observe from (2.19), we can obtain

$$f(y_k) \leq (1 - \gamma_k) f(y_{k-1}) + \gamma_k l_f(z_k, x_k) + \gamma_k \langle g_k, x - x_k \rangle$$
$$+ \frac{\beta_k \gamma_k}{2}\left[\|x_{k-1} - x\|^2 - \|x_k - x\|^2\right] + \eta_k \gamma_k$$
$$+ M\gamma_k \|x_k - x_{k-1}\| - \frac{\gamma_k \beta_k}{2}\|x_k - x_{k-1}\|^2$$
$$= (1 - \gamma_k) f(y_{k-1}) + \gamma_k l_f(z_k, x) + \gamma_k \langle \delta_k, x - x_k \rangle$$
$$+ \frac{\beta_k \gamma_k}{2}\left[\|x_{k-1} - x\|^2 - \|x_k - x\|^2\right] + \eta_k \gamma_k$$
$$+ M\gamma_k \|x_k - x_{k-1}\| - \frac{\gamma_k \beta_k}{2}\|x_k - x_{k-1}\|^2.$$

Using the above inequality and the fact that

$$\langle \delta_k, x - x_k \rangle + M\|x_k - x_{k-1}\| - \frac{\beta_k}{2}\|x_k - x_{k-1}\|^2$$
$$= \langle \delta_k, x - x_{k-1} \rangle + \langle \delta_k, x_{k-1} - x_k \rangle + M\|x_k - x_{k-1}\| - \frac{\beta_k}{2}\|x_k - x_{k-1}\|^2$$
$$\leq \langle \delta_k, x - x_{k-1} \rangle + \frac{\|\delta_k\|_*^2 + M^2}{\beta_k},$$

we obtain for all $x \in X$,

$$f(y_k) \leq (1 - \gamma_k) f(y_{k-1}) + \gamma_k f(x) + \eta_k \gamma_k + \frac{\beta_k \gamma_k}{2}\left[\|x_{k-1} - x\|^2 - \|x_k - x\|^2\right]$$
$$+ \gamma_k \langle \delta_k, x - x_{k-1} \rangle + \frac{\gamma_k (\|\delta_k\|_*^2 + M^2)}{\beta_k}.$$

Following the same procedure as we obtain (2.22) in Theorem 3, we conclude that

$$f(y_k) - f(x) \leq \frac{\beta_k \gamma_k}{2} D_X^2 + \Gamma_k \sum_{i=1}^{k} \frac{\gamma_i}{\Gamma_i}\left[\eta_i + \frac{\|\delta_i\|_*^2 + M^2}{\beta_i} + \langle \delta_i, x - x_{i-1} \rangle\right]. \quad (4.23)$$

Note that by our assumptions on SFO, the random variables $\delta_i$ are independent of the search point $x_{i-1}$ and hence $\mathbb{E}[\langle \delta_i, x^* - x_{i-1}\rangle] = 0$. In addition, relation (2.3) implies that $\mathbb{E}[\|\delta_i\|_*^2] \leq \sigma^2$. Using the previous two observations and taking expectation on both sides of (4.23) (with $x = x^*$) we obtain (4.22). The proof of Part b) and c) are the same as in Theorem 3. ∎

Now we provide a set of parameters $\{\beta_k\}, \{\gamma_k\}$, and $\{\eta_k\}$, which leads to an optimal complexity bound on the number of calls to the SFO oracle as well as a comparable complexity bound on $\text{LOsep}_X$.

**Corollary 15** *Assume that the outer iteration limit $N \geq 1$ is given. If $\{\beta_k\}, \{\gamma_k\}$, and $\{\eta_k\}$ in Algorithm 6 are set to*

$$\beta_k = \frac{\sqrt{N(\sigma^2+M^2)}}{D_X}, \quad \gamma_k = \tfrac{1}{k}, \quad \text{and} \quad \eta_k = \frac{D_X\sqrt{\sigma^2+M^2}}{\sqrt{N}}. \tag{4.24}$$

*Under assumptions (2.2) and (2.3), we have*

$$\mathbb{E}\left[f(y_N) - f(x^*)\right] \leq \frac{5 D_X \sqrt{\sigma^2+M^2}}{2\sqrt{N}}, \quad \forall k \geq 1. \tag{4.25}$$

*As a consequence, the total number of calls to the SFO and $\text{LOsep}_X$ oracles performed by Algorithm 6 for finding a stochastic $\epsilon$-solution of (1.1), respectively, can be bounded by*

$$\mathcal{O}\left\{\frac{(\sigma^2+M^2)D_X^2}{\epsilon^2}\right\}, \tag{4.26}$$

*and*

$$\mathcal{O}\left\{\log\left(\frac{(\sigma^4+M^4)D_X^4}{\Lambda\epsilon^4}\right) + \frac{(\sigma^4+M^4)D_X^4}{\epsilon^4}\right\} \text{ with probability } 1 - \Lambda. \tag{4.27}$$

**Proof** It can be easily seen from (4.24) that we have

$$\Gamma_k = \tfrac{1}{k}, \tag{4.28}$$

and hence

$$\frac{\beta_k \gamma_k}{\Gamma_k} = \frac{\sqrt{N(\sigma^2+M^2)}}{D_X},$$

which implies that (2.15) (or (2.17)) holds. Moreover, we have

$$\sum_{i=1}^N \frac{\eta_i \gamma_i}{\Gamma_i} \leq D_X\sqrt{N(\sigma^2+M^2)}, \quad \sum_{i=1}^N \frac{\gamma_i(\sigma^2+M^2)}{\Gamma_i \beta_i} \leq D_X\sqrt{N(\sigma^2+M^2)}.$$

Using the bound in (4.22), we obtain (4.25), which implies that the total number of outer iteration $N$ can be bounded by $\mathcal{O}\left((\sigma^2+M^2)D_X^2/\epsilon^2\right)$ under the assumptions (2.2) and (2.3). Therefore, the number of calls to the SFO oracle is bounded by $\mathcal{O}\left((\sigma^2+M^2)D_X^2/\epsilon^2\right)$, since we set the batch-size $B_k \equiv 1$.

Similarly, we can perform the same procedure as we obtain (2.29) in Corollary 4 to obtain a good estimation for $\Phi_0^k$ (cf. Line 1 in LCG procedure) at the $k$-th outer iteration. By Cauchy-Schwarz and triangle inequalities, we have with probability $1 - \Lambda$,

$$\Phi_0^k = \langle g_k - f'(z_k), x_{k-1} - x\rangle + \langle f'(z_k), x_{k-1} - x\rangle\}$$
$$\leq \left(\sqrt{\tfrac{N\sigma^2}{\Lambda}} + \|f'(z_k)\|_*\right) D_X \leq \left(\sqrt{\tfrac{N\sigma^2}{\Lambda}} + M\right) D_X, \tag{4.29}$$

where the last inequality follows from (4.21). Note that we always have $\eta_k < \alpha\beta_k D_X^2$. Therefore, it follows from the bound in (2.18), (4.24), and (4.29) that the total number of inner iterations can be bounded by

$$\sum_{k=1}^{N} T_k \leq \sum_{k=1}^{N} \left[ 4\alpha \left( \log \frac{\Phi_0^k}{\alpha\beta_k D_X^2} + 1 \right) + \log \frac{\Phi_0^k}{\eta_k} + \frac{8\alpha^2 \beta_k D_X^2}{\eta_k} + 2 \right]$$

$$\leq \sum_{k=1}^{N} \left[ 5\alpha \log \left( \frac{N}{\sqrt{\Lambda}} + \sqrt{N} \right) + 8\alpha^2 N \right] + (4\alpha + 2)N$$

$$= \mathcal{O}\left( \log \frac{N^2}{\Lambda} + N^2 + N \right),$$

which implies our bound in (4.27). ∎

Furthermore, if we can estimate the distance from the initial point to the set of optimal solutions, i.e., there exits an estimate $D_0$ s.t. $\|x_0 - x^*\| \leq D_0 \leq D_X$, the complexity bounds (4.26) and (4.27) can be improved slightly in terms of the dependence on $D_X$.

## 5. Experimental results

We present preliminary experimental results showing the performance of CALSGD compared to OFW for stochastic optimization. As examples we use the video co-localization problem, which can be solved by quadratic programming over a path polytope, different structured regression problems, and quadratic programming over the standard spectrahedron. In all cases we use objective functions of the form $\|Ax - b\|^2$, with $A \in \mathbb{R}^{m \times n}$, i.e., $m$ examples over a feasible region of dimension $n$. In each example there is a density parameter $d$ specifying the fraction of non-zero entries in $A$. We compute $b = Ax^*$ with some feasible point $x^*$ so that in all examples the optimal value is 0. For comparability we use a batch size of 128 for all algorithms to compute each gradient and the full matrix $A$ for the actual objective function values. Since the function evaluations are not used by any algorithm, each algorithm has only the information provided by the 128 examples sampled in that specific round. All graphs show the function value using a logscale on the vertical axis. We implemented all algorithms using `Python 2.7` using `Gurobi 7.0` Gurobi Optimization (2016) as the solver for our linear models. Note that our test instances are smooth problems, but OFW and CALSGD can also be applied to solve non-smooth problems.

In Figure 1 we compare the performance of three algorithms: CALSGD, stochastic conditional gradient sliding (SCGS) and OFW. As described above SCGS is the non-lazy counterpart of CALSGD. In the four graphs of Figure 1 we report the objective function value over the number of iterations, the wall clock time in seconds, the number of calls to the linear oracle, and the number of gradient evaluations in that order. In all these measures, our proposed algorithms outperform OFW by multiple orders of magnitude. As expected in number of iterations and number of gradient evaluations both versions CALSGD and SCGS perform equally well, however in wall clock time and in the number of calls to the linear oracle we observe the advantage of the weaker LOsep oracle over LO.

For the rest of the results we compare only the best version of our algorithm CALSGD with OFW. We report on each example the performance of the algorithms over the number of iterations and wall clock time in seconds. The three problem we consider are the following.

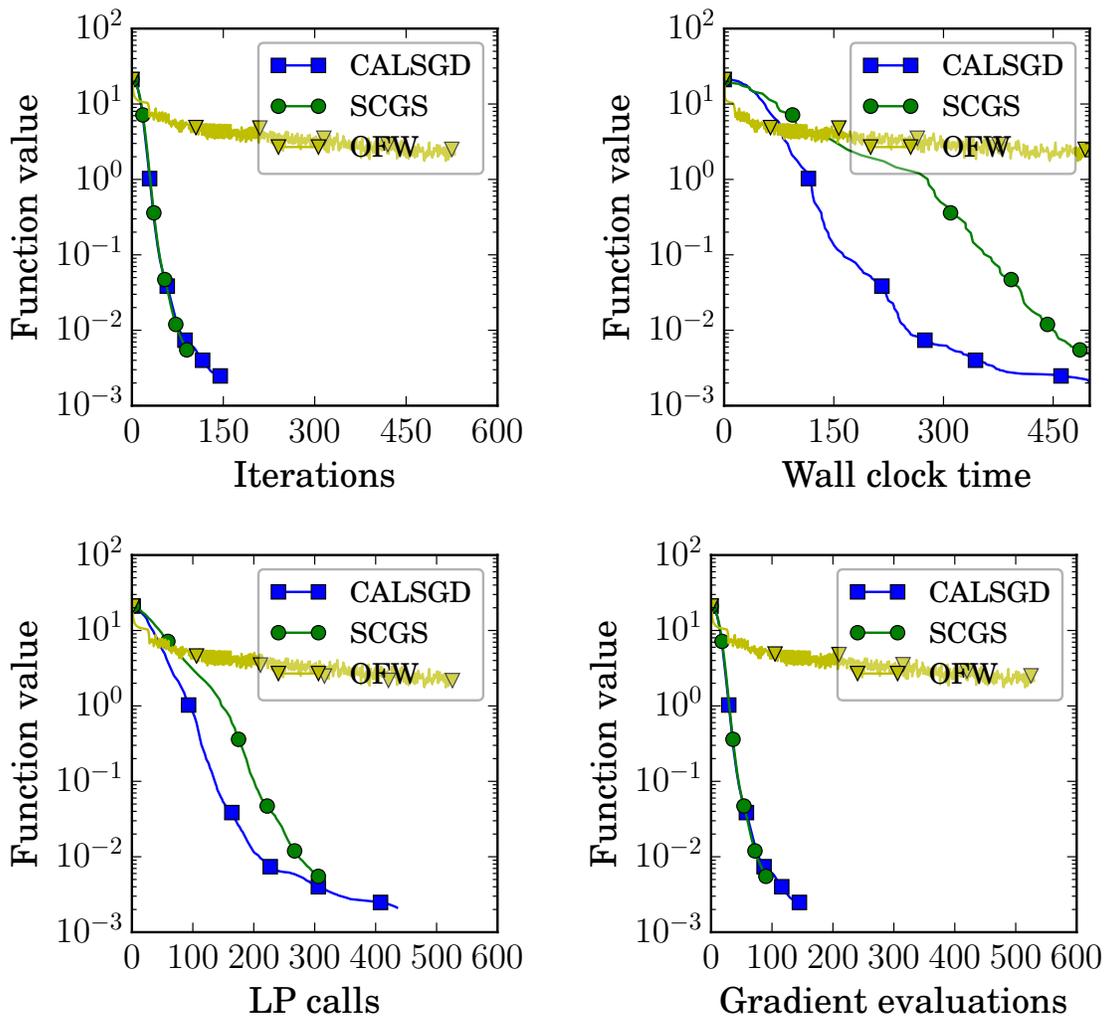

Figure 1: Performance of CALSGD and its non-lazy variant SCGS on a structured regression problem compared to OFW. The feasible region of this instance is the flow-based formulation of the convex hull of Hamiltonian cycles on 9 nodes and has dimension $n = 162$. Time limit is 500 seconds.

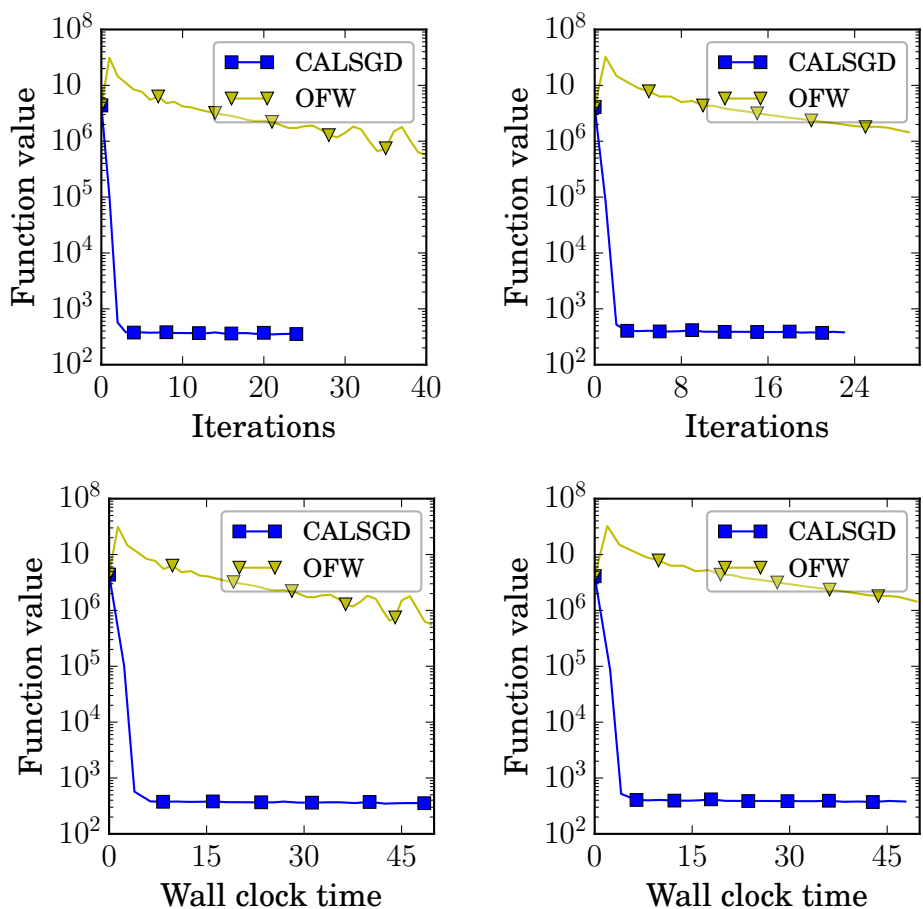

Figure 2: Two small video co-localization instances. On the left: `road_paths_01_DC_a` instance ($n = 29682$ and $m = 10000$). On the right: `road_paths_01_DC_b` instance ($n = 29682$ and $m = 10000$). We observe in both cases a significant difference in function value of multiple orders of magnitude after only a few seconds.

**Video co-localization** Video co-localization is the problem of identifying an object over multiple frames of a video. As shown by Joulin et al. (2014) this problem can be solved by quadratic programming over a path/flow polytope. In Figures 2, 3 and 4 we show that our algorithm CALSGD performs significantly better than OFW on this type of instances. We use path polytopes available at http://lime.cs.elte.hu/~kpeter/data/mcf/road/. The non-zero entries of $A$ in this section are chosen uniformly from $[0,1]$ and the density parameter we used is $d = 0.8$.

**Structured regression** For our structured regression instances we solve the objective function $\|Ax - b\|^2$ as described before over different polytopes. In Figure 5 the feasible region is the convex hull of all Hamiltonian cycles of graphs of different size. In Figure 6 the polytopes are the standard formulation of the cut problem and the Birkhoff polytope.

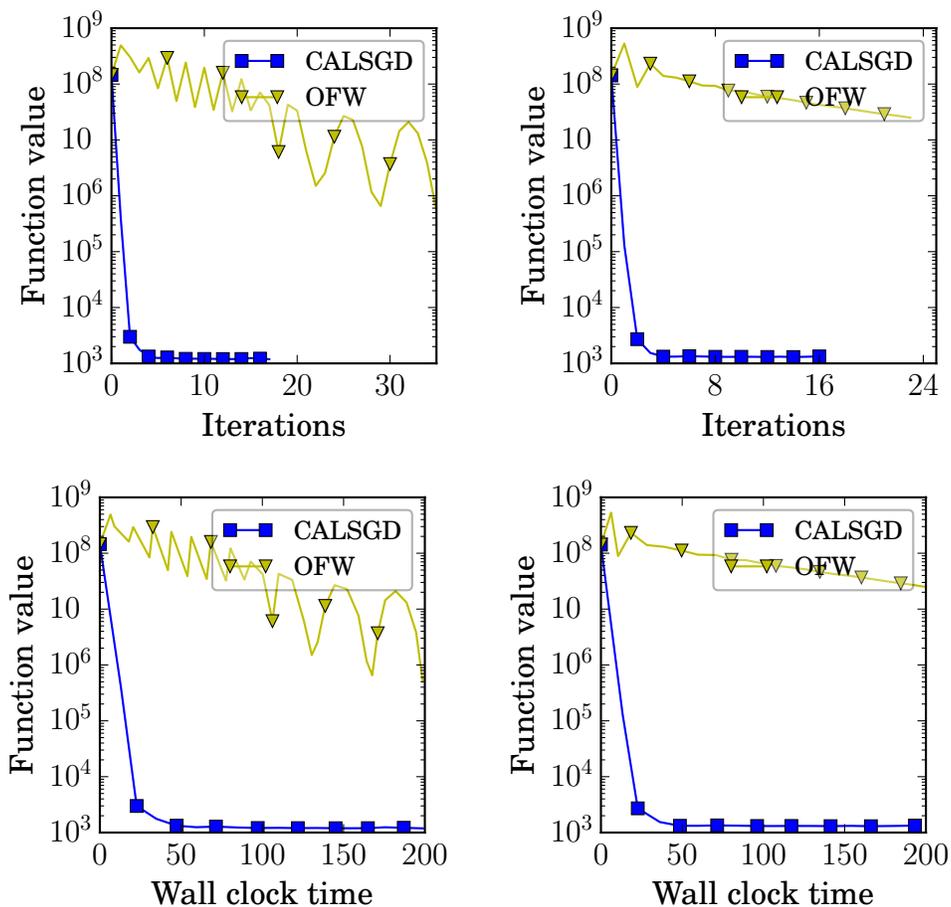

Figure 3: Two medium sized video co-localization instances. On the left: road_paths_02_DE_a instance ($n = 119520$ and $m = 10000$). On the right: road_paths_02_DE_b instance ($n = 119520$ and $m = 10000$). Similar results as in Figure 2: CALSGD achieves after a few seconds objective function values that OFW does not achieve in the whole time window of 200 seconds.

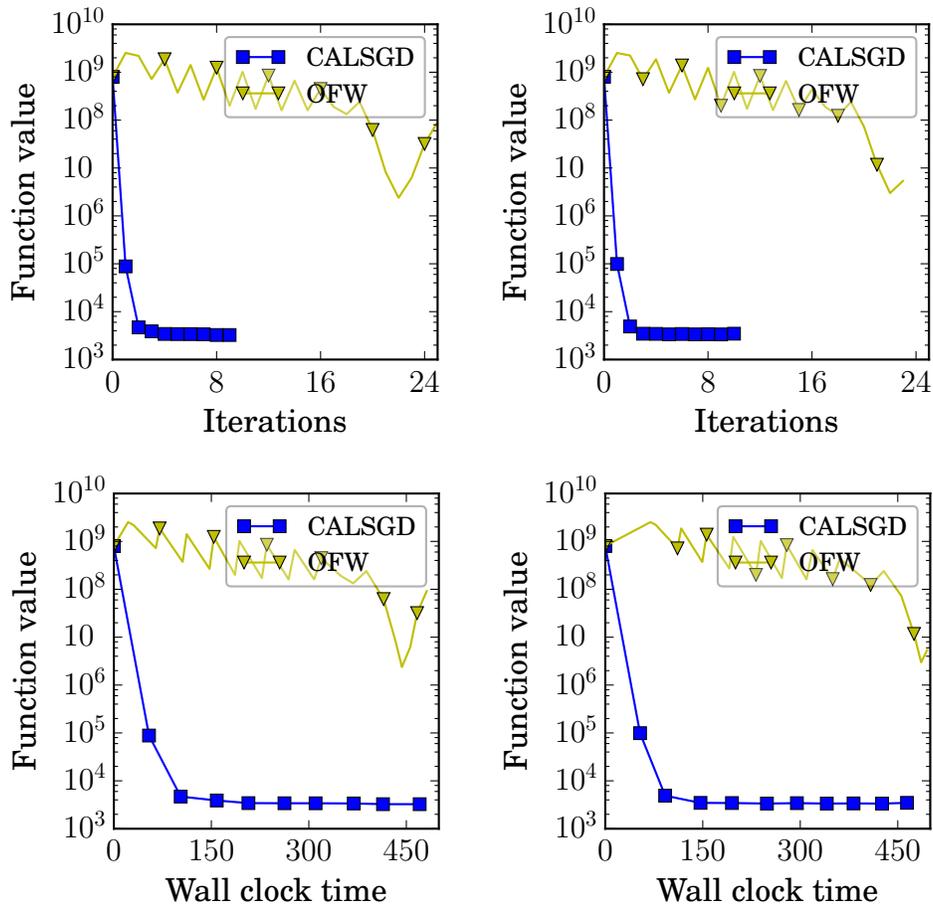

Figure 4: Two large video co-localization instances. On the left: `road_paths_03_NH_a` instance ($n = 262958$ and $m = 10000$). On the right: `road_paths_03_NH_b` instance ($n = 262958$ and $m = 10000$). CALSGD has a better performance in both, iterations and wall clock time.

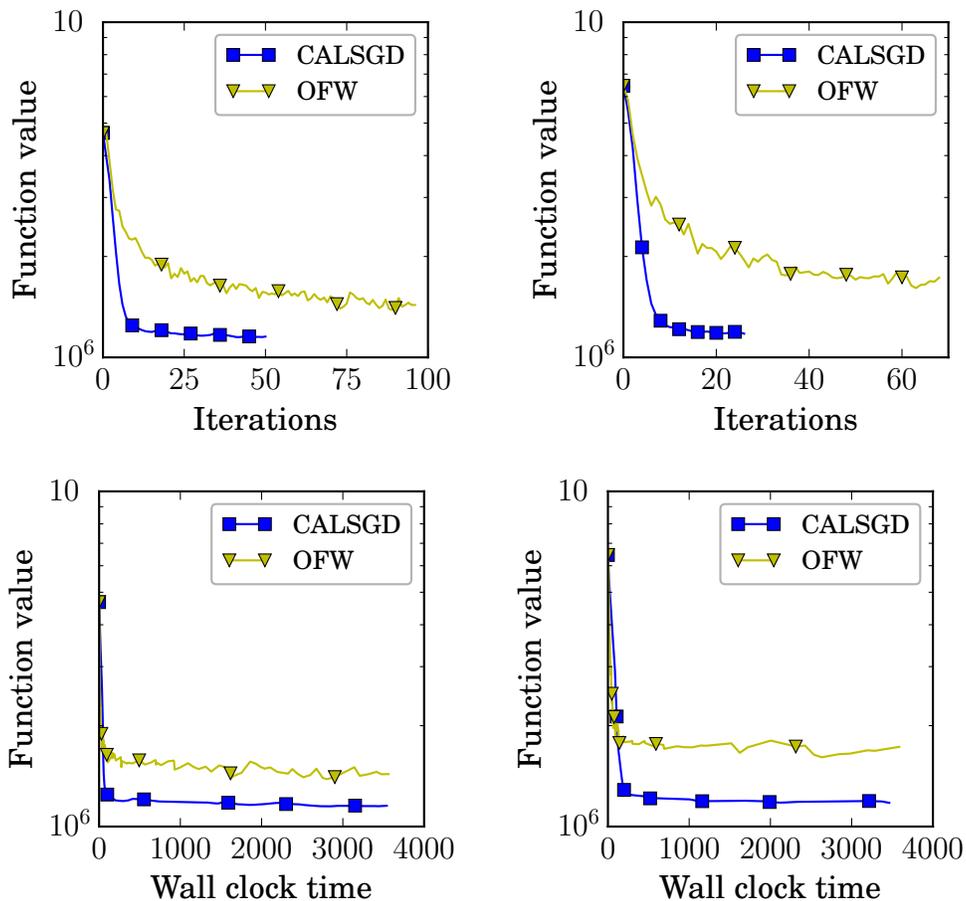

Figure 5: Structured regression problem over the convex hull of all Hamiltonian cycles of a graph on 11 nodes ($n = 242$) on the left and 12 nodes ($n = 288$) on the right. We used a density of $d = 0.6$ for $A$ and $m = 10000$. On both instances we can see that our proposed method CALSGD achieves lower values much faster, both in number of iterations as well as in wall clock time.

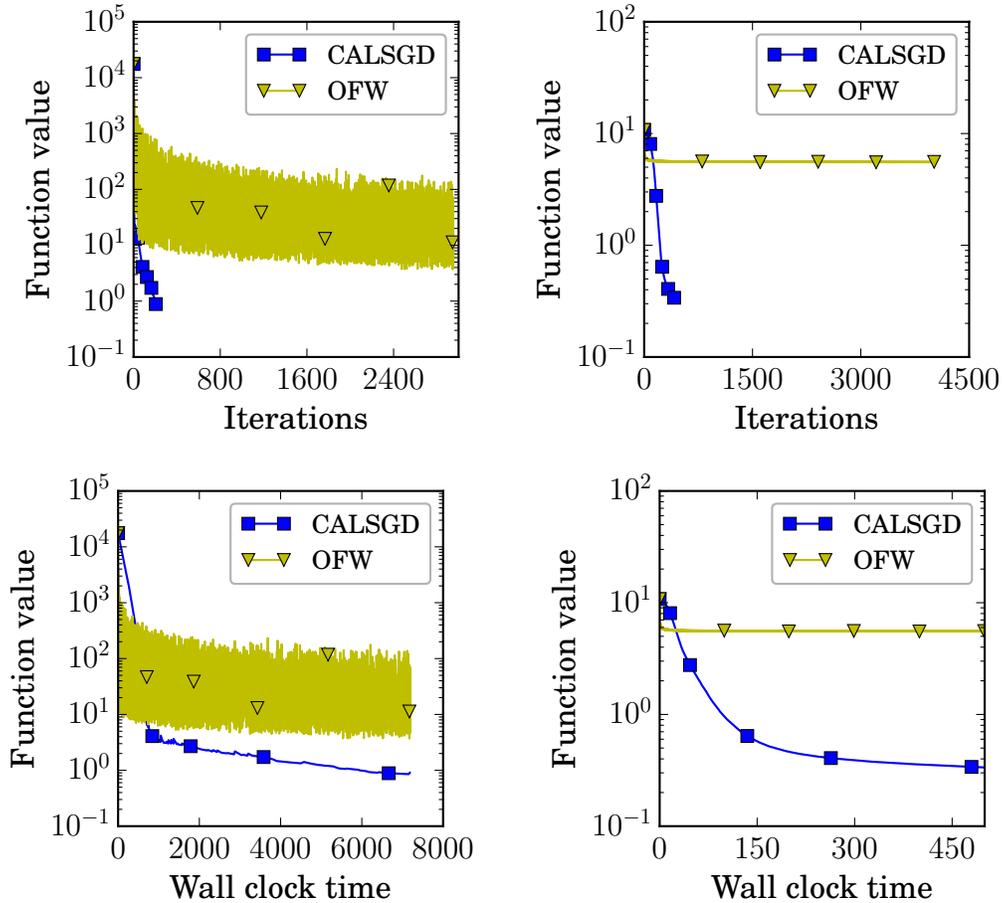

Figure 6: Structured regression problem over the cut polytope for a graph on 23 vertices on the left and the Birkhoff polytope containing all doubly stochastic matrices of size $100 \times 100$ on the right. In both cases we used $m = 10000$ rows for the matrix $A$, on the left a density of $d = 0.6$ and on the right $d = 0.8$. The number of iterations computed in the given time between CALSGD and OFW is quite significant, however in all test cases CALSGD achieves better function values in the smaller number of iterations. In the example of the Birkhoff polytope it almost looks like as if OFW converges suboptimally, however this is due to the large number of iterations required: the convergence rate of OFW as shown by Hazan and Kale (2012) is $\mathcal{O}(T^{-1/4})$, so if we compute the improvement with logarithmic scale, from, e.g., iteration 1500 to iteration 4500, we get $-1/4(\log(1500) - \log(4500)) \approx 0.12$ (the constants hidden in the $\mathcal{O}$-notation get canceled due to the logarithm and the difference) and therefore indeed fits to the observation on the graph.

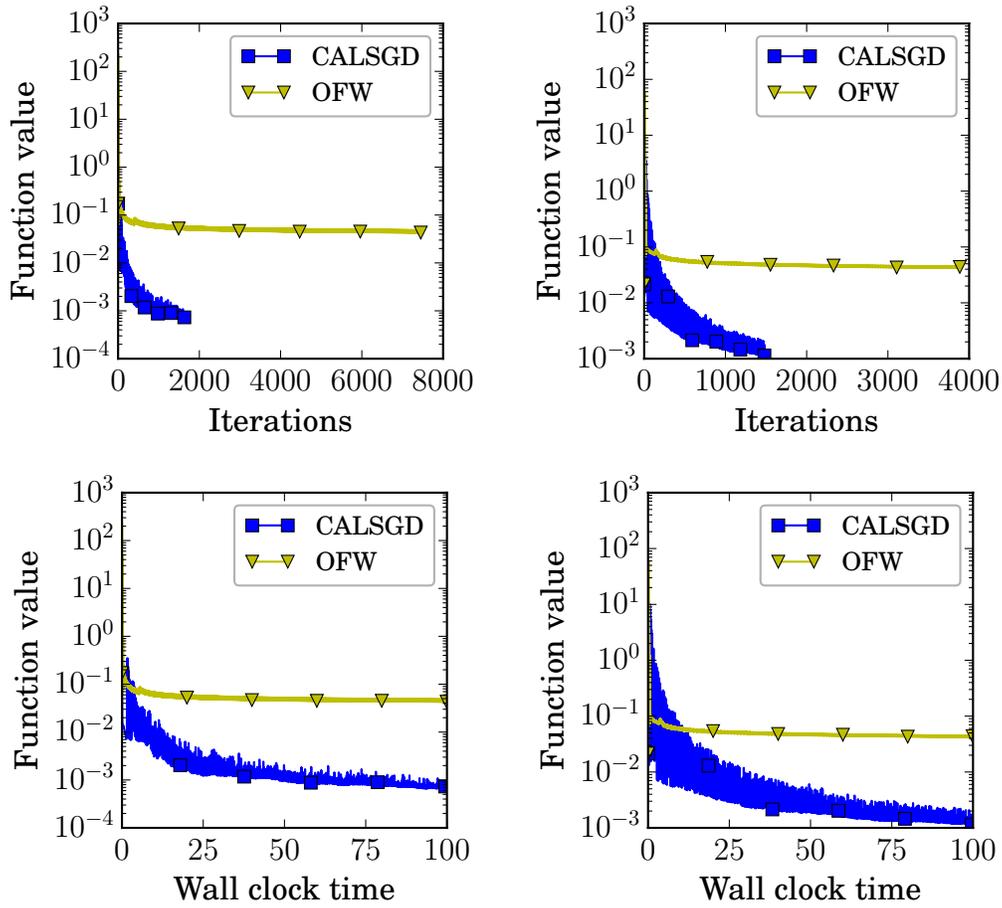

Figure 7: Quadratic optimization over the standard spectrahedron of size $n = 50$. On the left we use $m = 10000$ on the right $m = 20000$. In both cases CALSGD performs better than OFW both in iterations as well as in wall clock time. As described in Figure 6 the impression of suboptimal convergence of OFW can be explained by the very high number of iterations required.

**Convex optimization over spectrahedra** We consider instances of the problem of finding the minimum of a convex function over the standard spectrahedron, which is defined as $S_n := \{X \in \mathbb{R}^{n \times n} \mid X \succcurlyeq 0, tr(X) = 1\}$. In this case the linear minimization problem for an objective function $C$ is solved by computing an eigenvector for the largest eigenvalue of $-C$. We use the same method to implement $\text{LOsep}_{S_n}$. We show results on three different sized instances, in Figure 7 for $n = 50$, Figure 8 for $n = 100$ and in Figure 9 for $n = 150$.

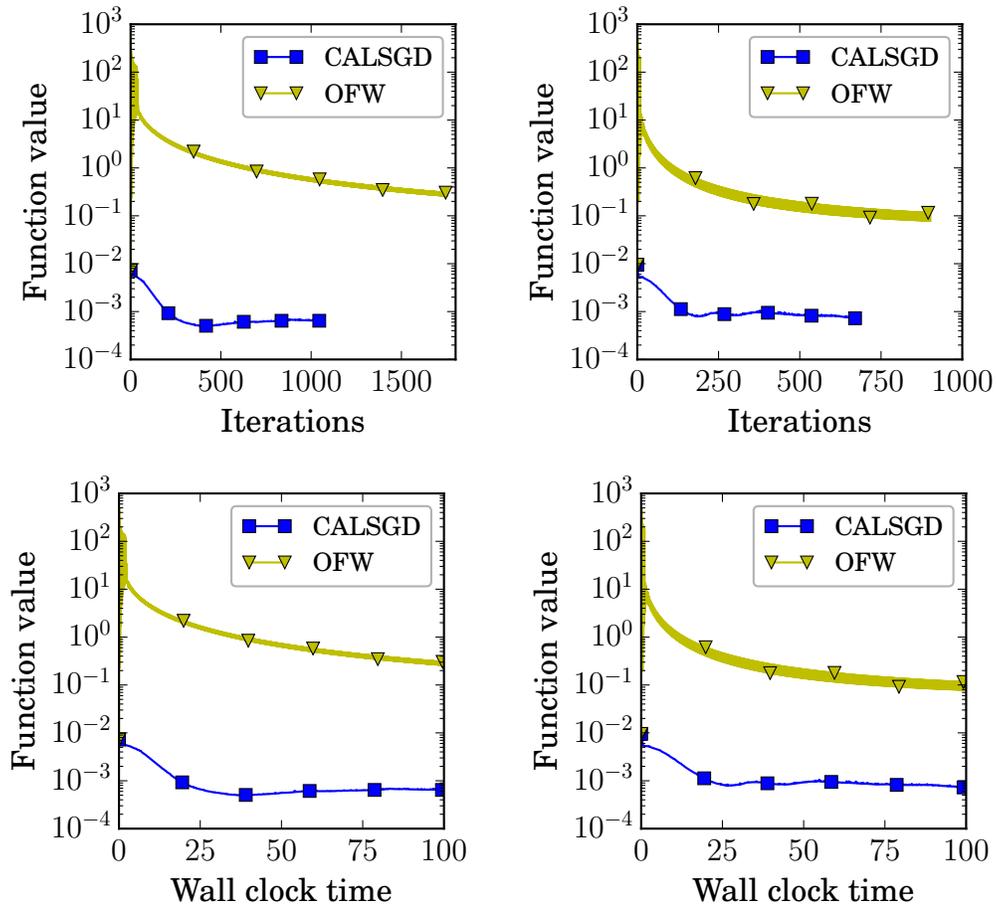

Figure 8: Medium sized quadratic optimization over the standard spectrahedron ($n = 100$). Again we chose $m = 10000$ on the left and $m = 20000$ on the right. The CALSGD method achieves values multiple orders of magnitude better within the given time window.

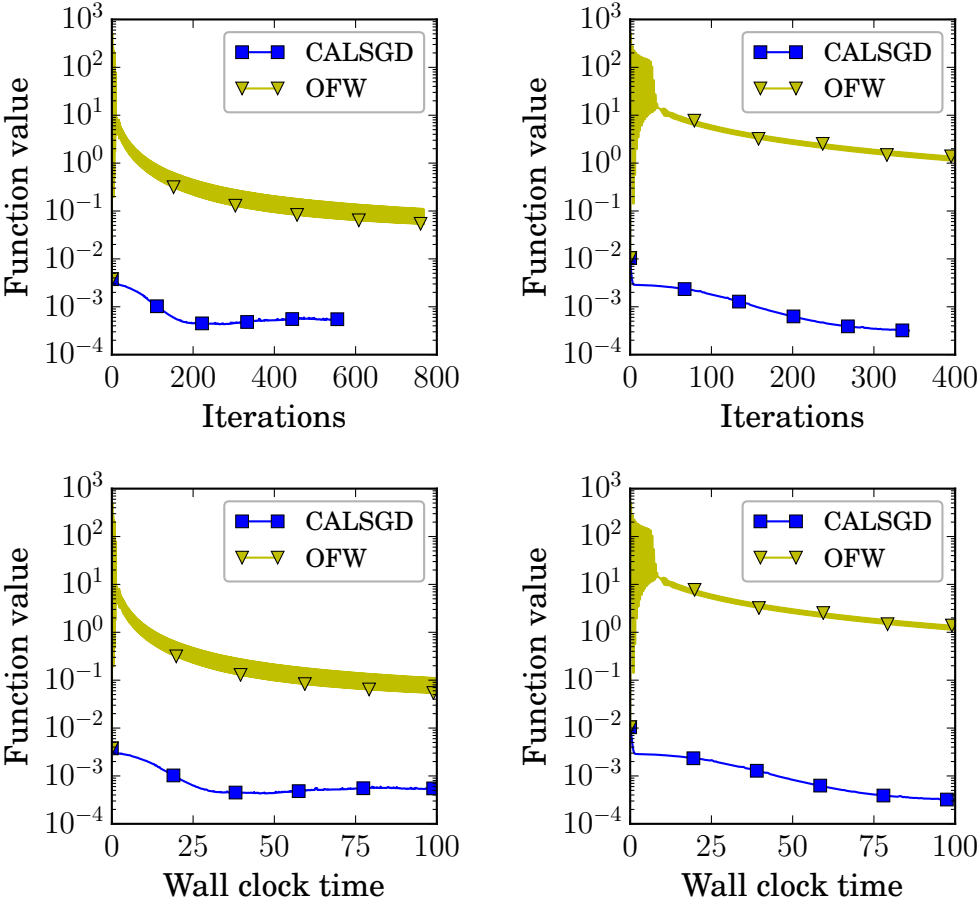

Figure 9: Large quadratic optimization over the standard spectrahedron ($n = 150$), with $m = 10000$ on the left and $m = 20000$ on the right. The behaviour and the achieved objective function values are very similar to the medium size instances in Figure 8.

## Acknowledgments

Research reported in this paper was partially supported by NSF CAREER award CMMI-1452463, as well as NSF grants 1637473 and 1637474, and Office of Naval Research grant N00014-16-1-2802.